\documentclass[10pt,twocolumn,letterpaper]{article}
\usepackage[pagenumbers]{cvpr}
\usepackage{graphicx}
\usepackage{amsmath}
\usepackage{amssymb}
\usepackage{booktabs}

\usepackage[pagebackref,breaklinks,colorlinks]{hyperref}
\usepackage{xcolor}
\usepackage{makecell} 

\usepackage{multicol}
\usepackage{multirow}
\usepackage{textcomp}

\usepackage{color}
\usepackage{colortbl}
\definecolor{Gray}{gray}{0.95}

\usepackage[capitalize]{cleveref}
\crefname{section}{Sec.}{Secs.}
\Crefname{section}{Section}{Sections}
\Crefname{table}{Table}{Tables}
\crefname{table}{Tab.}{Tabs.}

\begin{document}
\title{
Multi-View Partial (MVP) Point Cloud Challenge 2021 \\ on Completion and Registration: Methods and Results
}

\author{Liang Pan$^{1}$ \;
Tong Wu$^{2}$ \;
Zhongang Cai$^{1,3,4}$ \;
Ziwei Liu$^{1}$ \\
Xumin Yu$^{5}$ \; Yongming Rao$^{5}$ \; Jiwen Lu$^{5}$ \; Jie Zhou$^{5}$ \\
Mingye Xu$^{6,7}$ \; Xiaoyuan Luo$^{8}$ \; Kexue Fu$^{4,8}$ \; Peng Gao$^{4}$ \; Manning Wang$^{8}$ \; Yali Wang$^{7\ddagger}$ \; Yu Qiao$^{4,7\ddagger}$\\
Junsheng Zhou$^{9\star}$ \; Xin Wen$^{10\star}$ \; Peng Xiang$^{9}$ \; Yu-Shen Liu$^{9\ddagger}$ \; Zhizhong Han$^{11}$ \\
Yuanjie Yan$^{12}$ \; Junyi An$^{12}$\\ 
Lifa Zhu$^{13}$ \; Changwei Lin$^{13}$ \; Dongrui Liu$^{14}$ \; Xin Li$^{15}$ \; Francisco Gómez-Fernández$^{13}$\\
Qinlong Wang$^{16}$ \; Yang Yang$^{16}$\\
\fontsize{8}{8}\selectfont $^1$S-Lab, Nanyang Technological University \;
$^2$SenseTime-CUHK Joint Lab, The Chinese University of Hong Kong \;
$^3$Sensetime Research \\
\fontsize{8}{8}\selectfont $^4$Shanghai AI Laboratory \;
$^{5}$Department of Automation, Tsinghua University \; 
$^{6}$University of Chinese Academy of Sciences \\
\fontsize{8}{8}\selectfont $^{7}$ShenZhen Key Lab of Computer Vision and Pattern Recognition, SIAT-SenseTime Joint Lab,\\
\fontsize{8}{8}\selectfont Shenzhen Institutes of Advanced Technology, Chinese Academy of Sciences\\
\fontsize{8}{8}\selectfont $^{8}$Digital Medical Research Center, School of Basic Medical Science, Fudan University \; 
$^{9}$School of Software, BNRist, Tsinghua University \\
\fontsize{8}{8}\selectfont $^{10}$JD.com \; 
$^{11}$Wayne State University \;
$^{12}$State key Laboratory for novel software technology, Nanjing University \\
\fontsize{8}{8}\selectfont $^{13}$DeepGlint \;
$^{14}$Shanghai Jiao Tong University \; 
$^{15}$Sichuan University \; 
$^{16}$Xi'an Jiaotong University \\
}

\maketitle
\begin{abstract}
As real-scanned point clouds are mostly partial due to occlusions and viewpoints, reconstructing complete 3D shapes based on incomplete observations becomes a fundamental problem for computer vision.
With a single incomplete point cloud, it becomes the partial point cloud completion problem.
Given multiple different observations, 3D reconstruction can be addressed by performing partial-to-partial point cloud registration.
Recently, a large-scale Multi-View Partial (\textbf{MVP}) point cloud dataset has been released, which consists of over 100,000 high-quality virtual-scanned partial point clouds.
Based on the MVP dataset, this paper reports methods and results in the \textbf{Multi-View Partial Point Cloud Challenge 2021 on Completion and Registration}.
In total, 128 participants registered for the competition, and 31 teams made valid submissions.
The top-ranked solutions will be analyzed, and then we will discuss future research directions.
\end{abstract}
\vspace{-3mm}
\section{Introduction}
3D reconstruction for point clouds has been extensively explored in past decades.
Thanks to the rapid development of deep learning recently, many researchers study learning-based approaches to perform single-view completion~\cite{yuan2018pcn,pan2020ecg,pan2021variational,zhang2021unsupervised,yu2021pointr} and multi-view registration~\cite{wang2019prnet,li2019iterative,yuan2020deepgmr, fu2021robust,huang2021predator,pan2021robust,zhu2021point} for high-quality 3D reconstruction.
However, completion and registration for partial point clouds are far from being fully resolved by existing methods.

Recently, we have been established a versatile multi-view partial (MVP) point cloud dataset~\cite{pan2021variational}, 
which contains over 100,000 high-quality virtual-scanned partial point clouds and complete point clouds.
Employing the MVP~\cite{pan2021variational} dataset, we organized the \textit{Multi-View Partial Point Cloud Challenge 2021 on Completion and Registration} (MVP Challenge\footnote{Challenge website:\href{https://competitions.codalab.org/competitions/33430}{https://competitions.codalab.org/competitions/33430}}) collocated with the Workshop on Sensing, Understanding and Synthesizing Humans at ICCV2021\footnote{Workshop website: \href{https://sense-human.github.io/}{https://sense-human.github.io/}}.
The MVP Challenge lasted for nine weeks, from Jul. 12th, 2021 to Sep. 12th, 2021.
The goal of this challenge is to boost research on point cloud completion and registration.
A total of 128 participants registered for our competition, and 31 teams made valid submissions.
All participants are restricted to training their models using our prepared training data only for fair comparisons.
On Oct. 18th, 2021, the top-3 ranked approaches for each track are selected and rewarded.

In the following sections, we will introduce the completion track (Sec.~\ref{sec:com}) and the registration track (Sec.~\ref{sec:reg}) of the MVP challenge.
For each track, we will describe the settings, analyze the top-ranked solutions, and discuss potential future research directions.

\section{Single-View Partial Point Cloud Completion} \label{sec:com}
\paragraph{Overview.}
Given a partial observation, point cloud completion targets at reconstructing its complete 3D shape.
After registering for the MVP challenge, each team is able to submit their completion results for evaluation on the Codalab platform. 
All models are required to be trained by using our prepared training data only, and those trained models that achieve the best performance on the test data set can probably provide high-quality completion results on the extra-test data set.
We highlight that no additional strategies, such as pre-training, are allowed.

\paragraph{Dataset.}
The MVP Challenge 2021 on Point Cloud Completion mainly employs the MVP dataset~\cite{pan2021variational} that we proposed in CVPR 2021.
The MVP dataset is a large-scale multi-view partial point cloud dataset containing over 100,000 high-quality scans, which renders partial 3D shapes from 26 uniformly distributed camera poses for each 3D CAD model.
It provides a Training set with 62,400 partial-complete point cloud pairs and a Test set with 41,800 pairs.
Besides, we generate an extra-test set consisting of 59,800 partial-complete point cloud pairs following the same fashion, which is used for evaluating different completion methods in this challenge.
We suggest that future research works should only use the Test set for evaluation instead of the extra-test set.
Notably, each partial and/or complete point cloud has 2,048 points.

\paragraph{Evaluation Metric.}
Considering the computation efficiency, we use the symmetric Chamfer Distance (CD) Loss for evaluating completion methods.
Formally, the CD loss $\mathcal{L}_{CD}$ can be formulated as:
\begin{equation}
    \mathcal{L}_{\mathbf{CD}}(\mathbf{P}, \mathbf{Q}) = \frac{1}{|\mathbf{P}|}\sum_{x\in\mathbf{P}}\underset{y\in\mathbf{Q}}{\min}\|x-y\|^2 + \frac{1}{|\mathbf{Q}|}\sum_{y\in\mathbf{Q}}\underset{x\in\mathbf{P}}{\min}\|x-y\|^2,
\end{equation}
where $x$ and $y$ denote points that belong to two point clouds $\mathbf{P}$ and $\mathbf{Q}$, respectively.

\paragraph{Results.}
The benchmark results are reported in Table~\ref{tab:all_comp}.
In the following subsections, we are going to summarize and report their methods and experiments according to their submitted reports.

\begin{table}[]
\centering
\setlength{\tabcolsep}{13pt}
\caption{Top team results in MVP Completion Challenge 2021.}
\begin{tabular}{l|c|cc}
\toprule[1.5pt]
Method & Ranking & CD ($\times 10^4$) \\ 
\midrule\midrule
SPTNet & 3 & 5.15 \\
TSPCC & 2 & 5.01 \\
PointTr++ & 1 & 5.01 \\
\bottomrule[1.5pt]
\end{tabular}
\label{tab:all_comp}
\end{table}

\subsection{Solution of First Place}
\noindent{\textit{PoinTr++: Enhanced Geometry-Aware Transformers with Iterative Refinement}}

\noindent\textit{\\Team Members: Xumin Yu, Yongming Rao, Jiwen Lu, and Jie Zhou}

\noindent\textbf{\textit{\\General Method Description}}

\noindent Overall, the champion team uses PoinTr~\cite{yu2021pointr} to complete a point cloud from a partial input (shown in Fig.~\ref{figc1:c11}). Then they use multiple refinement blocks to iteratively denoise the prediction to produce the final point cloud.
\begin{itemize}
\item[$\bullet$] \textbf{Concatenation + refinement pipeline:} Many previous works like PCN~\cite{yuan2018pcn} and TopNet~\cite{tchapmi2019topnet} adopt a pipeline that is more similar to reconstruction, encoding the input point cloud as a single feature and reconstructing the completed point cloud by a decoder like FoldingNet~\cite{yang2018foldingnet}. While PoinTr~\cite{yu2021pointr} adopts the concatenation strategy, where the final prediction comes from the concatenation results of the inputs and the outputs of the model. The reconstruction pipeline struggles to keep the details of the input point clouds while the concatenation pipeline is facing the problem that the final prediction may be discontinuous in appearance. They propose a pipeline that combines these two strategies. They add reconstruction modules to the concatenated point clouds to further refine the results and make the final point cloud smooth and continuous. To keep the details of the input, they only predict the position shift vector for each point in the refinement block. In the newly proposed pipeline, they effectively combine the advantages of the concatenation-based and reconstruction-based methods.

\item[$\bullet$]\textbf{Iterative refinement:} They further investigate the refinement strategy and propose the iterative refinement method, which iteratively refines the predicted point clouds using several refinement blocks. They concatenate the origin input and the prediction point cloud from the previous step and send it into the next refinement block.

\begin{figure*}[]
  \centering
  \includegraphics[width = 0.8\linewidth]{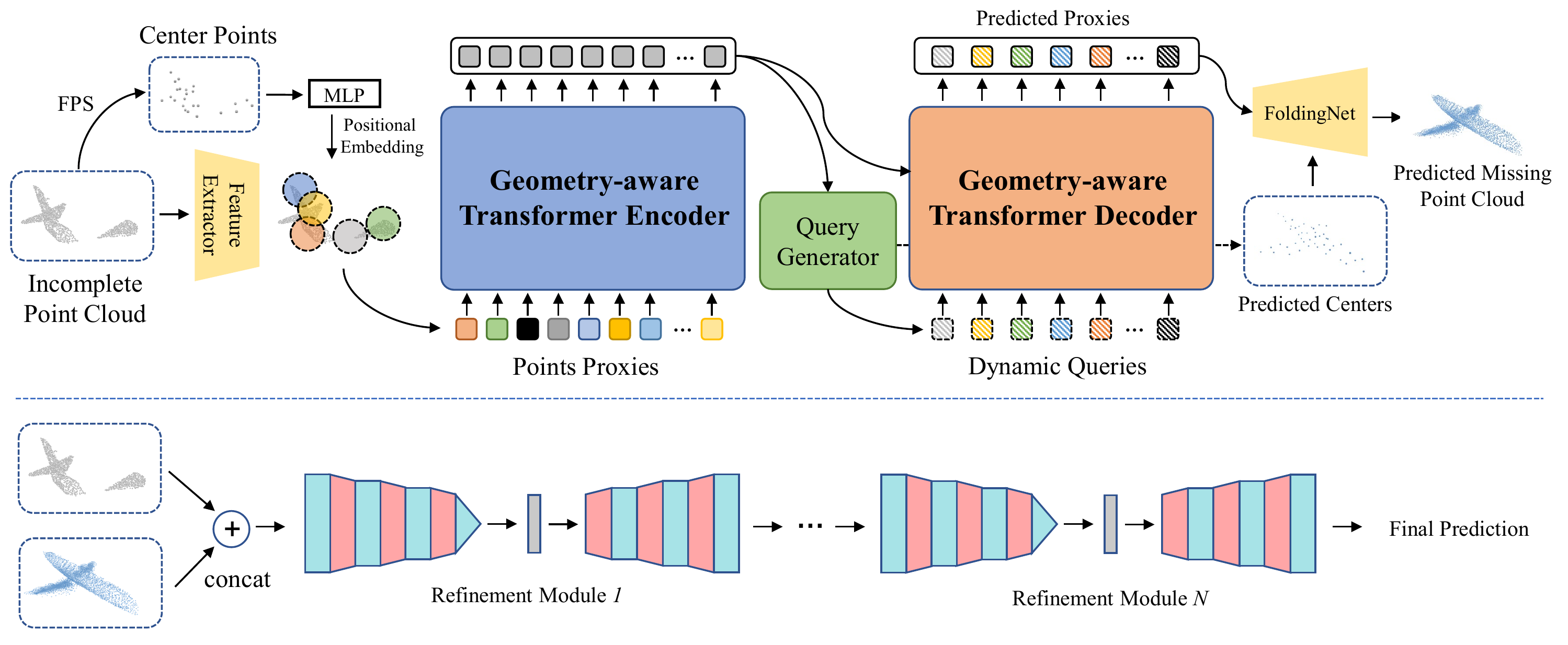}
  \vspace{-3mm}
  \caption{\small The pipeline of our PoinTr++. They use PoinTr~\cite{yu2021pointr} to complete a point cloud from a partial input. Then they use refinement blocks to iteratively denoise the prediction point cloud.}
  \label{figc1:c11}
\end{figure*}

A straightforward way to implement the iterative refinement is to add several refinement modules after PoinTr and optimize them in an end-to-end manner. However, it will cause two problems: 1) deeper model is harder to optimizer; 2) the end-to-end training brings a heavy computational cost and GPU memory consumption.
Therefore, they propose to iteratively add refinement blocks after the base model during the training. When adding one more refinement block, they freeze the model before it and only train the newly added module, which can save the most computation costs and decompose the optimization problem into several simpler sub-problems.

In their experiments, they select RENet proposed in the VRCNet paper~\cite{pan2021variational} as their refinement block because of its high performance. The module takes the concatenated point clouds as inputs and refine the point cloud using a hierarchical encoder-decoder architecture with Edge-preserved Pooling (EP) and Edge-preserved Unpooling (EU) modules which effectively learn multi-scale structural relations. 
\end{itemize}

\begin{table}[]
\small
\setlength{\tabcolsep}{13pt}
\caption{Comparisons of performance with existing methods on the MVP validation set. Both the input and output contain 2048 points. PoinTr+ use a single refinement module to produce the final prediction. CD loss multiplied by $10^4$.} 
\centering
\label{tabc1:c11}
\begin{tabular}{l | c c }
\toprule[1.5pt]
Model  &  CD & F1-Score@1\%\\
\midrule\midrule
PCN~\cite{yuan2018pcn}   & 9.77  & 0.32 \\
TopNet~\cite{tchapmi2019topnet} & 10.11 & 0.308 \\
CRN~\cite{wang2020cascaded}   & 7.25  & 0.434 \\
ECG~\cite{pan2020ecg}   & 6.64  & 0.476 \\
VRCNet~\cite{pan2021variational} & 5.96  & 0.499 \\
\hline
\rowcolor{Gray}   PoinTr~\cite{yu2021pointr} & 6.15  & 0.456 \\
\rowcolor{Gray}   PoinTr+ & 5.13 & 0.511 \\
\rowcolor{Gray}   PoinTr++ & \textbf{4.93} & \textbf{0.525} \\
\bottomrule[1.5pt]
\end{tabular}
\end{table}

\noindent\textbf{\textit{\\Training Description}}

\noindent The training process in the concatenation-refinement pipeline is multi-stage. They iteratively train each refinement module during the training phase. In our experiments on MVP benchmark, they add two refinement blocks after PoinTr, so a two-stage training is used.

In the first stage of the training phase, they jointly train a refinement module and PoinTr~\cite{yu2021pointr}. They set the learning rate to $1\times10^{-3}$, batch size to 32, and weight decay to $5\times10^{-2}$. The hidden dimension for PoinTr is set to 384. During the training, they use AdamW optimizer with WarmingUpCosLR scheduler to optimize the model. The L1 chamfer distance (CD-$\ell_1$) is adopted as our training loss. Specifically, they calculate CD-$\ell_1$ between four predicted point clouds (includes three coarse-grained predictions and 
one fine-grained prediction) and ground-truth. They adaptively adjust the weights for these four loss terms to obtain the final weighted loss during the training phase. In the first 10 epochs, they use the weights of [1, 1, 0.5, 0.1]. In the second 10 epochs, they use the weights of [1, 1, 1, 0.5]. For the remaining epochs of training, they use the weights of [1, 1, 1, 1].
They normalize the total loss by dividing the sum of the weights. They use an early-stop strategy in the first stage of training.
\footnote{They stop the training when the network converges on the test set.}

In the second phase, they add a refinement block after the first one. The weights of PoinTr and the first refinement block come from the first stage, and they initialize the additional refinement block with the weights of the first refinement block. In this phase of training, they freeze the PoinTr and the first refinement block. They set the batch size to 16 in this phase. All other hyper-parameters are the same as the first phase.

They stop adding more refinement blocks to the model since the performance will not be significantly improved on the MVP benchmark. 
Therefore, they only use two refinement modules for the sake of efficiency.

\noindent\textbf{\textit{\\Testing Description}}

\noindent In the test phase, they follow the standard procedure of point cloud completion. 
They send a partial point cloud containing 2048 points to the trained model and obtain a completed point cloud with 2048 points.
Results comparing against the other applications are reported in Table~\ref{tabc1:c11}.

\begin{figure*}[]
	\centering
	\includegraphics[width=0.8\linewidth]{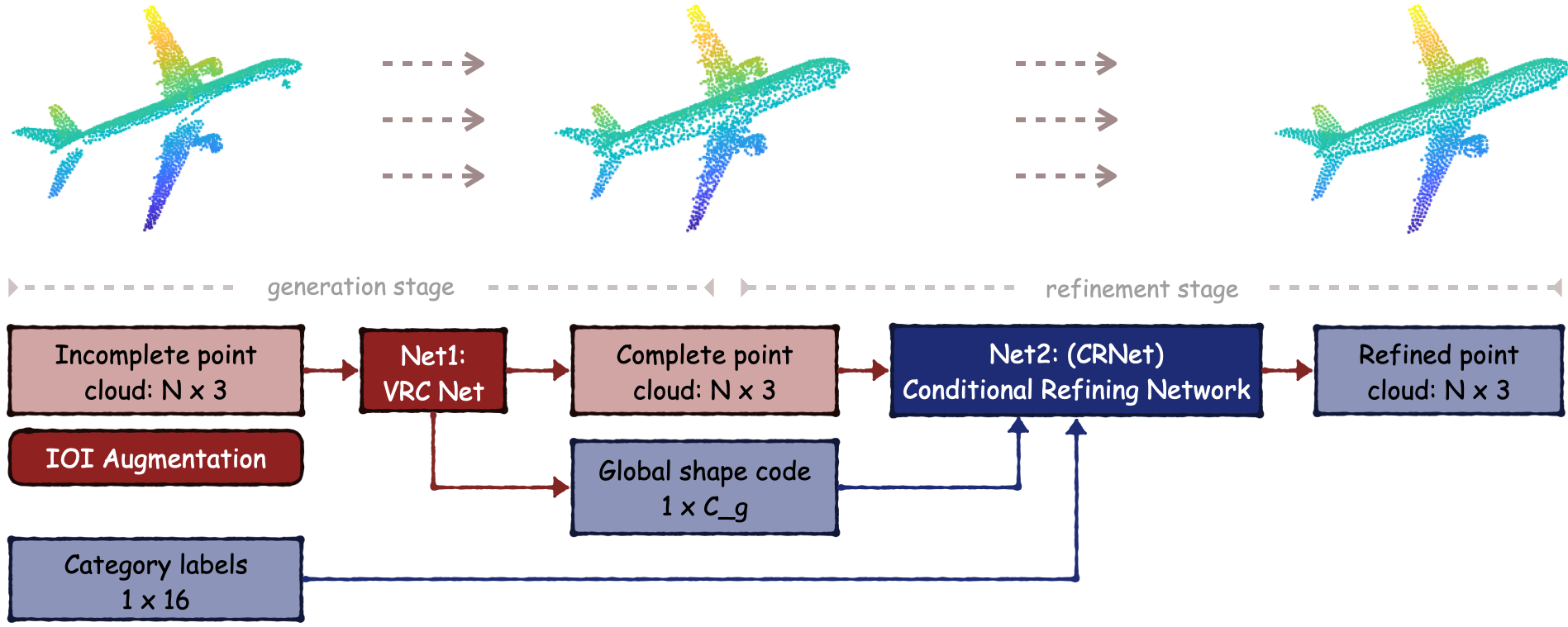}
	\vspace{-3mm}
	\caption{The overall network architecture.  The generation stage is used to generate the complete point cloud with the robustness of diversified incomplete structures. Subsequently, the refinement stage is used to refine the complete point clouds with discriminative underlying attributes of category label and global representation.}
	\label{figc2:c21}
\end{figure*}

\subsection{Solution of Second Place}
\noindent{\textit{Robust and Discriminative Two-Stage Point Cloud Completion with Semantic Refinement and IOI augmentation}}

\noindent{\textit{\\Team Members: Mingye Xu, Xiaoyuan Luo, Kexue Fu, Peng Gao, Manning Wang, Yali Wang$^\ddagger$, and Yu Qiao$^\ddagger$.}}

\noindent $^\ddagger$ denotes corresponding author

\begin{figure}[]
	\centering
	\includegraphics[width=1\linewidth]{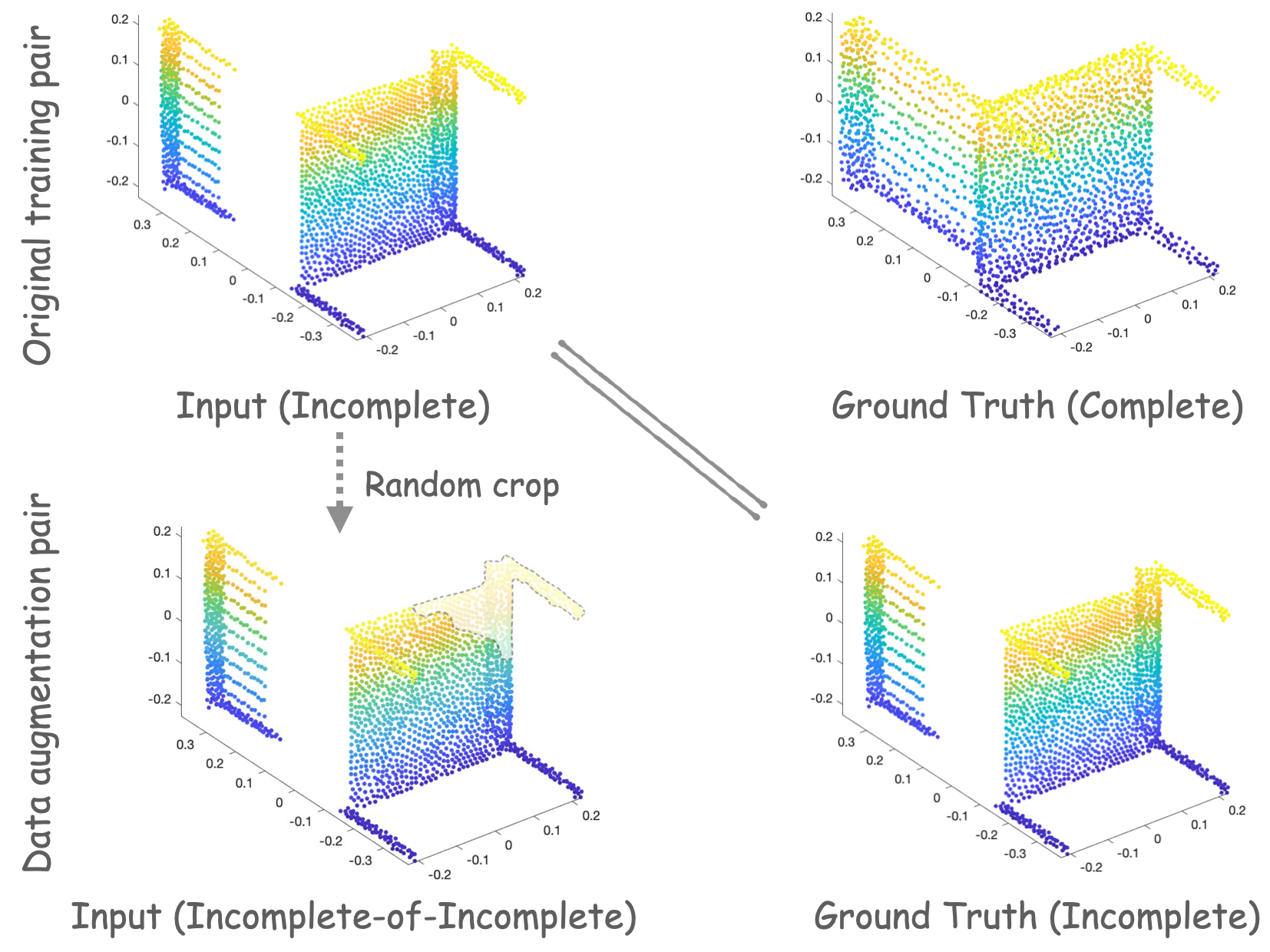}
	\caption{Incompletion-of-Incompletion data augmentation. They crop the original training  incomplete point clouds randomly as the new incomplete input, and treat the original incomplete point cloud as the ground truth.}
	\label{figc2:c22}
\end{figure}
\noindent\textbf{\textit{\\General Method Description}}

\noindent Fig. \ref{figc2:c21} illustrates the structure of their two-stage point cloud completion network, which consists of generation stage (Net1: VRCNet \cite{pan2021variational} with IOI data augmentation) and refinement stage (Net2: their Conditional Refining Network). 
The IOI data augmentation can increase the diversity of the incomplete point clouds to boost the generality of the generation network.
Then they use their Condition Refining Network (CRNet) to make more detailed refinement with the aid of semantic category information and shape codes.

\begin{figure}[]
	\centering
	\includegraphics[width=1\linewidth]{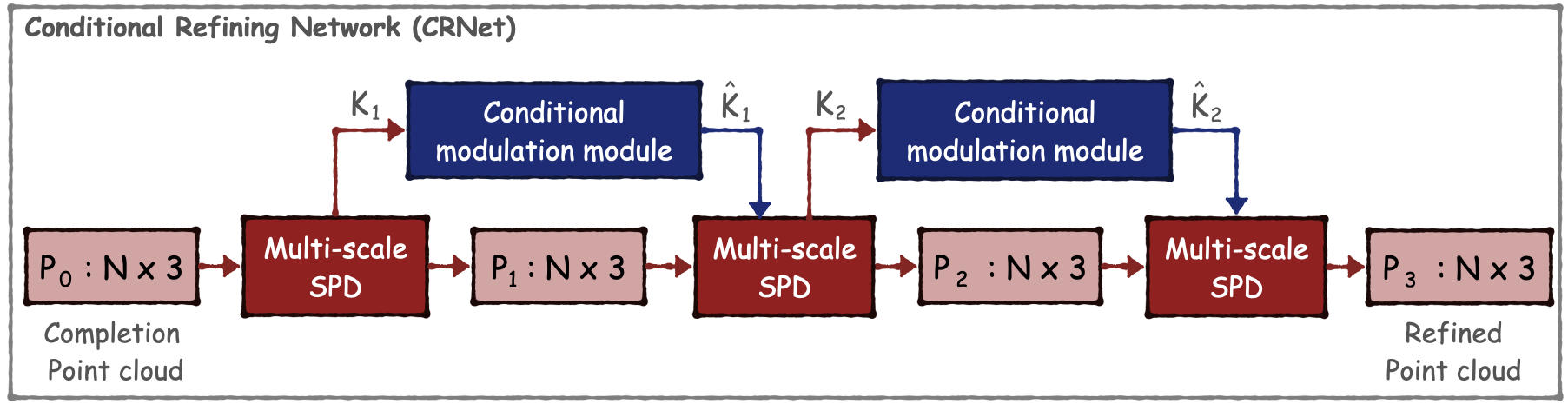}
	\caption{Their Conditional Refining Network includes three consecutive multi-scale SPD modules, and use conditional modulation module to adjust the displacement features.}
	\label{figc2:c23}
\end{figure}

\paragraph{Robust Point Cloud Generation: VRCNet with IOI Augmentation.} 
Their generation network is built upon the VRCNet \cite{pan2021variational} which consists of two consecutive encoder-decoder sub-networks that serve as “probabilistic modeling” (PMNet) and “relational enhancement” (RENet).
PMNet embeds global shape representation and latent distributions from the partial inputs and generates the coarse skeletons. 
Then RENet strives to enhance structural relations by learning multi-scale local point features, and reconstruct the fine complete point clouds on coarse skeletons.

\begin{itemize}
\item[$\bullet$] \textbf{IOI (Incompletion-of-Incompletion) Augmentation:}
To increase the robustness of point cloud generation,
they propose a novel Incompletion-of-Incompletion (IOI) data augmentation method.
As Figure~\ref{figc2:c22} shows, they can crop the incomplete point cloud randomly and feed it to the model to reconstruct the original incomplete point cloud.
This augmentation is aiming at increasing the diversity of global features and latent distributions from PMNet and makes the RENet more generalization capacity to variations of incomplete structures.
Verified by their experiments in Table~\ref{tabc2:analysis}, the proposed data augmentation can indeed improve the performance of the complement network.

\item[$\bullet$] \textbf{Self-Supervised Pretraining by Point Cloud Reconstruction:}
They also investigate the pre-training mechanism.
The self-supervised reconstruction pre-training can reach a good initial point across downstream fine-tuning completion tasks.
In this way, it can lead to wider optima and is easier to optimize compared with training from scratch. 
This can also improve the method performance, which is shown in Table \ref{tabc2:analysis}. 
\end{itemize}

\begin{figure}[]
	\centering
	\includegraphics[width=1\linewidth]{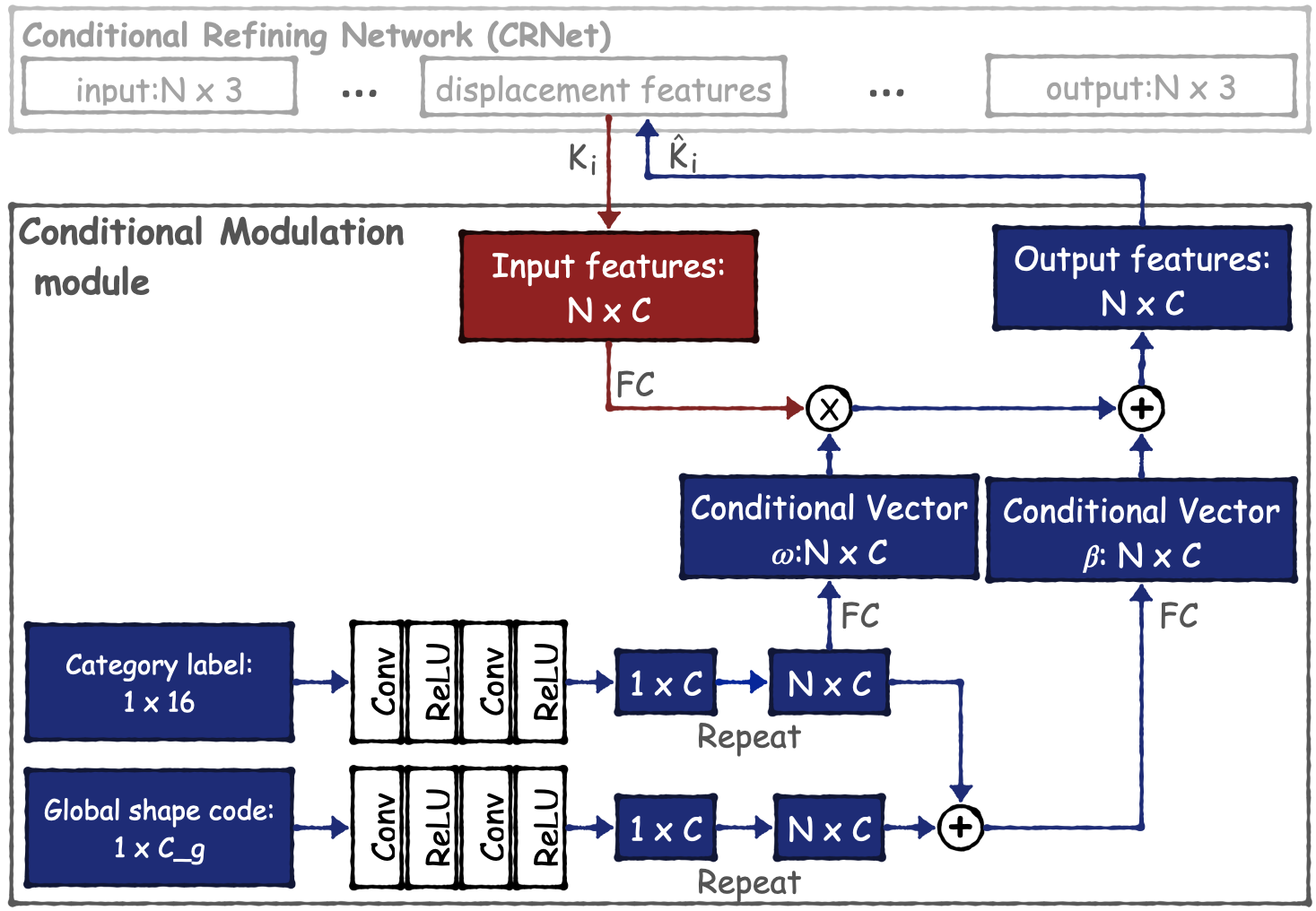}
	\caption{Conditional Modulation module.}
	\label{figc2:c24}
\end{figure}

\paragraph{Discriminative Point Cloud Refinement: Condition Refining Network with Semantic Guidance.}
Their refining network is aiming at refining the complete point cloud with more geometry details and more semantic information. 
Figure~\ref{figc2:c23} indicates the structure of their Conditional Refining network (CRNet), 
where
Conditional Modulation module can effectively adjust point-wise representation with semantic guidance,
while
Multi-Scale SPD module can refine the point cloud to show more geometrical structures with multi-scale context aggregation.
Details will be described below.

\begin{figure}[]
	\centering
	\includegraphics[width=1\linewidth]{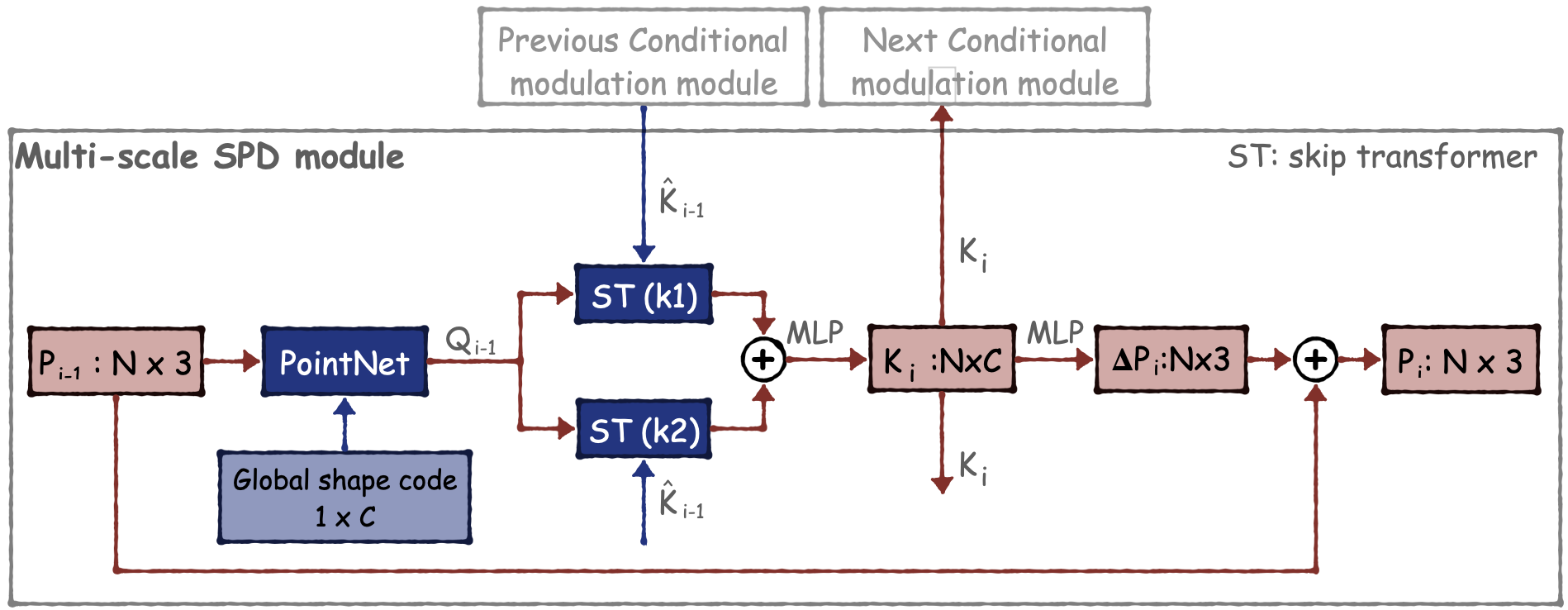}
	\caption{Multi-scale SPD Module with multi-scale skip transformers.}
	\label{figc2:c25}
\end{figure}
\begin{table}
	\setlength{\tabcolsep}{8pt}
	\caption{Shape completion results (CD loss multiplied by $10^{4}$) with various resolutions on the MVP dataset(2,048 points).}
	\centering
	\begin{tabular}{l|cc}
		\toprule[1.5pt]  { Method } &  { CD } &  { F1-Score@1\% } \\
		\midrule\midrule  { PCN \cite{yuan2018pcn}} & 9.77 & 0.320 \\
		{ TopNet \cite{tchapmi2019topnet}} & 10.11 & 0.308 \\
		{ MSN\cite{liu2020morphing}} & 7.90 & 0.432 \\
		{ Wang et. al.\cite{wang2020cascaded} } & 7.25 & 0.434 \\
		{ ECG\cite{pan2020ecg}} & 6.64 & 0.476 \\
		VRCNet \cite{pan2021variational} & {5.96} & {0.499} \\ \hline
		\rowcolor{Gray} \textbf{{CRNet} }& \textbf{5.27} &  \textbf{0.535}\\
		\bottomrule[1.5pt]
	\end{tabular}
	
	\label{tabc2:compair_others}
\end{table}
\begin{table}
    \caption{Shape completion results (CD loss multiplied by $10^{4}$) with various resolutions on the MVP dataset(16,384 points).}
    \setlength{\tabcolsep}{8pt}
	\centering
	\begin{tabular}{l|cc}
		\toprule[1.5pt]  { Method } &  { CD } &  { F1-Score@1\% } \\
		\midrule\midrule  { PCN \cite{yuan2018pcn}} & 6.02 & 0.638 \\
		{ TopNet \cite{tchapmi2019topnet}} & 6.36 & 0.601 \\
		{ MSN\cite{liu2020morphing}} & 4.90 & 0.710 \\
		{ Wang et. al.\cite{wang2020cascaded} } & 4.30 & 0.740 \\
		{ ECG\cite{pan2020ecg}} & 3.58 & 0.753 \\
		{GRNet \cite{xie2020grnet}} & 3.87 & 0.692 \\
		{NSFA \cite{zhang2020detail}} & 3.77 & 0.783 \\
		VRCNet \cite{pan2021variational} & {3.02} & {0.796} \\ \hline
		\rowcolor{Gray} \textbf{{CRNet} }&\textbf{2.51}& \textbf{0.824}\\
		\bottomrule[1.5pt]
	\end{tabular}
	
	\label{tabc2:compair_others_16384}
\end{table}

\begin{itemize}
    \item[$\bullet$] \textbf{Conditional Modulation Module:} The utilization of underlying shape attributes (global shape codes and semantic category information) can encourage the local representation closer to the global discrimination of the same object, which can be applied as the guidance of the point cloud refinement.
    Existing methods only merge the global information through the concatenation with the local representation, however the concatenation is not effective enough and it largely increases the weight of MLPs (Model F in Table \ref{tabc2:analysis}).
    These methods also ignore the important category information which contains discriminative semantics.
    To this end, they propose a lightweight Conditional Modulation Module for point cloud refinement.
    In addition to achieving the adjustment on global point cloud representation, the proposed module can be easily extended to learn local enhancement effects for point cloud refining.

    \begin{table*}
\setlength{\tabcolsep}{4.5pt}
\caption{Ablation studies of our method on MVP completion challenge. Training data:  MVP training set; VAL data: MVP testing set; Public Test: MVP Extra-Test set. (CD  loss multiplied by $10^4$) }
\vspace{-3mm}
\small
	\centering
	\begin{tabular}{cc|cc|c|cc}
		\toprule[1.5pt]
        Model & Based model  & Generation (Net1) & Refinement (Net2) & Strategy description & CD (Test) & CD (Extra-Test) \\
        \midrule\midrule
       \fontsize{7.5}{7.5}\selectfont  {A(baseline)} & \fontsize{7.5}{7.5}\selectfont - & \fontsize{7.5}{7.5}\selectfont VRCNet & \fontsize{7.5}{7.5}\selectfont - & \fontsize{7.5}{7.5}\selectfont - & \fontsize{7.5}{7.5}\selectfont 5.96 & \fontsize{7.5}{7.5}\selectfont 6.08\\ 
       \fontsize{7.5}{7.5}\selectfont {B} & \fontsize{7.5}{7.5}\selectfont {A} & \fontsize{7.5}{7.5}\selectfont VRCNet & \fontsize{7.5}{7.5}\selectfont - & \fontsize{7.5}{7.5}\selectfont Self-supervised pre-training & \fontsize{7.5}{7.5}\selectfont 5.78 (0.18$\downarrow$) & \fontsize{7.5}{7.5}\selectfont 5.91 (0.17$\downarrow$)  \\ 
       \fontsize{7.5}{7.5}\selectfont  {C} & \fontsize{7.5}{7.5}\selectfont {A} & \fontsize{7.5}{7.5}\selectfont VRCNet & \fontsize{7.5}{7.5}\selectfont - & \fontsize{7.5}{7.5}\selectfont IOI augmentation & \fontsize{7.5}{7.5}\selectfont 5.83 (0.13$\downarrow$)  & \fontsize{7.5}{7.5}\selectfont 5.93 (0.15$\downarrow$) \\ 
        {D} & \fontsize{7.5}{7.5}\selectfont {C} & \fontsize{7.5}{7.5}\selectfont VRCNet & \fontsize{7.5}{7.5}\selectfont Spatial Refiner \cite{li2021point} & \fontsize{7.5}{7.5}\selectfont Add refining module & \fontsize{7.5}{7.5}\selectfont 5.66 (0.20$\downarrow$)  & \fontsize{7.5}{7.5}\selectfont 5.76 (0.17$\downarrow$) \\ 
        {E} & \fontsize{7.5}{7.5}\selectfont {C} & \fontsize{7.5}{7.5}\selectfont VRCNet & \fontsize{7.5}{7.5}\selectfont SPD \cite{xiang2021snowflakenet} & \fontsize{7.5}{7.5}\selectfont Add refining module & \fontsize{7.5}{7.5}\selectfont 5.41 (0.42$\downarrow$)  & \fontsize{7.5}{7.5}\selectfont 5.50 (0.43$\downarrow$) \\ 
        {F} & \fontsize{7.5}{7.5}\selectfont {E} & \fontsize{7.5}{7.5}\selectfont VRCNet & \fontsize{7.5}{7.5}\selectfont CRNet (SPD) 
     & \fontsize{7.5}{7.5}\selectfont  Concatenate shape codes & \fontsize{7.5}{7.5}\selectfont 5.41 (0.00$\downarrow$)  & \fontsize{7.5}{7.5}\selectfont 5.50 (0.00$\downarrow$)  \\
         
        {G} & \fontsize{7.5}{7.5}\selectfont {E} & \fontsize{7.5}{7.5}\selectfont VRCNet & \fontsize{7.5}{7.5}\selectfont CRNet (SPD)
     & \fontsize{7.5}{7.5}\selectfont Add Conditional Modulation module & \fontsize{7.5}{7.5}\selectfont 5.32  (0.09$\downarrow$)  & \fontsize{7.5}{7.5}\selectfont 5.41 (0.09$\downarrow$) \\ 
       {H} & \fontsize{7.5}{7.5}\selectfont  {G} & \fontsize{7.5}{7.5}\selectfont VRCNet & \fontsize{7.5}{7.5}\selectfont CRNet (Multi-scale SPD)
     & \fontsize{7.5}{7.5}\selectfont Add multi-scale skip transformers & \fontsize{7.5}{7.5}\selectfont 5.27 (0.05$\downarrow$)   & \fontsize{7.5}{7.5}\selectfont 5.35 (0.06$\downarrow$)  \\
     \bottomrule[1.5pt]
	\end{tabular}
	
	\label{tabc2:analysis}
\end{table*}
    \begin{table*}
\setlength{\tabcolsep}{8.75pt}
\caption{Ablation studies of our method on MVP completion challenge. Training data: MVP training set and test set; Public Test: MVP extra-test set. (CD  loss multiplied by $10^4$)  }
\vspace{-3mm}
\small
	\centering
	\begin{tabular}{cc|cc|c|c}
	\toprule[1.5pt]
    Model & Based model  & Generation (Net1) & Refinement (Net2) & Strategy description & CD (Public Test) \\
    \midrule\midrule
   \fontsize{7.5}{7.5}\selectfont {A(baseline)} & \fontsize{7.5}{7.5}\selectfont - & \fontsize{7.5}{7.5}\selectfont VRCNet & \fontsize{7.5}{7.5}\selectfont - & \fontsize{7.5}{7.5}\selectfont - & \fontsize{7.5}{7.5}\selectfont   5.79 \\
   \fontsize{7.5}{7.5}\selectfont {C} & \fontsize{7.5}{7.5}\selectfont {}{A} & \fontsize{7.5}{7.5}\selectfont VRCNet & \fontsize{7.5}{7.5}\selectfont - & \fontsize{7.5}{7.5}\selectfont IOI augmentation & \fontsize{7.5}{7.5}\selectfont   5.61  (0.18$\downarrow$)  \\
   \fontsize{7.5}{7.5}\selectfont {D} & \fontsize{7.5}{7.5}\selectfont {}{C} & \fontsize{7.5}{7.5}\selectfont VRCNet & \fontsize{7.5}{7.5}\selectfont Spatial Refiner \cite{li2021point} & \fontsize{7.5}{7.5}\selectfont Add refining module & \fontsize{7.5}{7.5}\selectfont 5.58 (0.03$\downarrow$) \\
    \fontsize{7.5}{7.5}\selectfont {E} & \fontsize{7.5}{7.5}\selectfont {}{C} & \fontsize{7.5}{7.5}\selectfont VRCNet & \fontsize{7.5}{7.5}\selectfont SPD \cite{xiang2021snowflakenet} & \fontsize{7.5}{7.5}\selectfont Add refining module  & \fontsize{7.5}{7.5}\selectfont 5.33 (0.28$\downarrow$)  \\
   \fontsize{7.5}{7.5}\selectfont {G} & \fontsize{7.5}{7.5}\selectfont {}{E} & \fontsize{7.5}{7.5}\selectfont VRCNet & \fontsize{7.5}{7.5}\selectfont CRNet (SPD)
 & \fontsize{7.5}{7.5}\selectfont Conditional modulation module & \fontsize{7.5}{7.5}\selectfont   5.06  (0.27$\downarrow$) \\ 
     \fontsize{7.5}{7.5}\selectfont {H} & \fontsize{7.5}{7.5}\selectfont  {}{G} & \fontsize{7.5}{7.5}\selectfont VRCNet & \fontsize{7.5}{7.5}\selectfont CRNet (Multi-scale SPD)
 & \fontsize{7.5}{7.5}\selectfont Add multi-scale skip transformers  & \fontsize{7.5}{7.5}\selectfont   5.01 (0.05$\downarrow$)  \\ 
    \bottomrule[1.5pt]
	\end{tabular}
	
	\label{tabc2:analysis2}
\end{table*}

    As Figure~\ref{figc2:c25} shows, to enable the network to have the ability of handling operations that require semantic category information and global shape codes, they modulate the intermediate displacement features of the CRNet as follows:
    \begin{equation}
        \mathcal{\hat{K}}_{i}= \sigma(\mathcal{{K}}_{i}) \odot {w} + \beta,
    \end{equation}
    where $\odot$ denotes the element-wise multiplication operation and $\sigma$ is MLP layers, $\mathcal{{K}}_{i}\in \mathrm{R}^{N \times C}$ is the intermediate displacement features from Conditional Refining Network, 
    ${w}, \beta \in \mathrm{R}^{1 \times C}$ are affine parameters that are estimated from the point cloud category labels $ S \in \mathrm{R}^{1 \times 16}$ and point cloud global codes $ G \in \mathrm{R}^{1 \times C}$ from the previous generation network:
    \begin{equation}
        w = \gamma(S),
    \end{equation}
    \begin{equation}
        \beta = \gamma(S) + \theta(G),
    \end{equation}
    where $\gamma$ and $\theta$ are all MLP layers.
    
    They use conditional vector $w$ to affect the cluster centers of the local representation and use conditional vector $\beta$ to fine-tune the variance in the feature space.  Thus, they can achieve point feature global adjustment with only a few parameters. 
    The local features are encouraged to be more closer of the same object than features of other objects, such that the local representations of each object can be affected by the distinct semantic information and global shape codes.
    Therefore, the model is not easy to be confused with similar local structures under different semantic information.
    
    \item[$\bullet$] \textbf{Multi-Scale SPD module:} To reveal fine local geometric details on the complete shape, existing methods\cite{wang2020cascaded, yuan2018pcn,zhang2020detail} usually adopt folding-based strategy \cite{yang2018foldingnet} to obtain the variations for learning different displacements for the duplicated points. However, the folding-based strategy ignores the local shape characteristics contained in the original point due to the same 2D grids for sampling. Different from the folding based strategy, SnowflakeNet \cite{xiang2021snowflakenet} use SPD (Snowflake Point Deconvolution) to reformulate the generation of child points from parent points as a growing process of snowflake, where the shape characteristic embedded by the parent point features is extracted and inherited into the child points through a point-wise splitting operation. 
    They also introduce a novel skip-transformer \cite{xiang2021snowflakenet} to learn splitting patterns in SPD module which can learn shape context and spatial relationship between child points and parent points.

    \begin{figure*}[]
    \centering
    \includegraphics[width=1\linewidth]{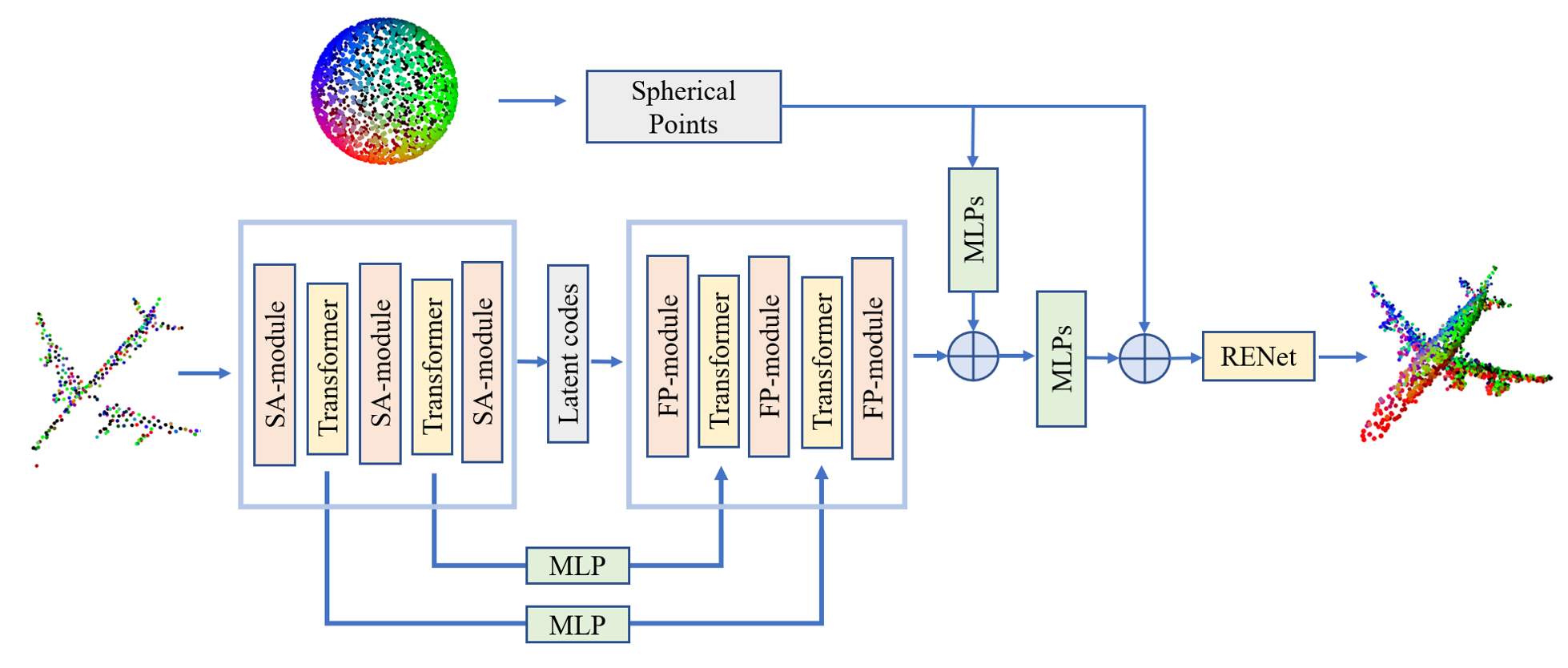}
    \vspace{-7mm}
    \caption{The framework of SPTNet.}
    \label{figc3:model}
\end{figure*}
    
    Their CRNet aims to refine the local geometric details on the complete point cloud.
    Inspired by SnowflakeNet \cite{xiang2021snowflakenet}, they use a similar structure as SPD.
    Different from SPD \cite{xiang2021snowflakenet}, their input is predicted completion point cloud with $ N = 2048$ points from Net1, and they did not use the point-wise splitting operation to increase the number of points. In contrast, they only obtain the variations of coordinates on each point in their multi-scale SPD module as Fig. \ref{figc2:c25} shows. 
    In order to progressively refine the local geometric details, three multi-scale SPDs are used in their Conditional Refining network, which is shown in Fig. \ref{figc2:c23}. 
    To facilitate consecutive multi-scale SPDs to refine points in a coherent manner, they use a skip-transformer to learn and refine the spatial context from different layers.
    Moreover, to improve the robustness to the variations on the diversity of local structures, they apply multi-scale skip transformers with different local regions in their multi-scale SPD module.
    
    As Fig. \ref{figc2:c25} shows, in the $i$-th multi-scale SPD module, they take the refined point cloud  from previous layer as $P_{i-1} \in \mathrm{R}^{N \times 3}$, and they extract the per-point feature $\mathcal{Q}_{i-1}  $ from $P_{i-1} \in \mathrm{R}^{N \times 3}$ and global shape codes  by  the basic PointNet \cite{qi2017pointnet}. 
    Then they send the displacement feature $\mathcal{\hat{K}}_{i-1} $ from the previous conditional modulation module and $\mathcal{Q}_{i-1} $ into two skip transformers with different local regions for local feature learning.
    Then they fed the multi-scale local features to MLP, and obtained the displacement feature $\mathcal{K}_{i}$ of the current layer.
    They can use $\mathcal{K}_{i}$ to generate the point displacement $\Delta \mathcal{P}_{i}$:
    \begin{equation}
    	\Delta \mathcal{P}_{i}=\tanh \left(\operatorname{MLP}\left(\mathcal{K}_{i}\right)\right),
    \end{equation}
    where $\tanh$ is hyper-tangent activation.
    Finally, the output point cloud is update as:
    \begin{equation}
    	\mathcal{P}_{i}={\mathcal{P}}_{i-1}+\Delta \mathcal{P}_{i}.
    \end{equation}
    
\end{itemize}

\noindent\textbf{\textit{\\Training description}}
\begin{itemize}
    \item[$\bullet$] \textbf{Stage one}: 
    They train the generation network with their IOI augmentation.  
    The training strategy is same as VRCNet \cite{pan2021variational}.  The batch size is set to 64.  A total of 150 epochs were executed.  The optimizer is  Adam optimizer.
    The training settings are similar with VRCNet \cite{pan2021variational}.
    \item[$\bullet$] \textbf{Stage two}: 
    They add the refining network (their proposed CRNet) and remove the data augmentation. The learning rate is initialized as $ 1 \times 10^{-4}$. 
    The batch size is set to 64.  A total of 70 epochs were executed. 
    In this stage, they only train the refining network , and the generation part (Net1 in Fig. \ref{figc2:c21}) of the model is fixed after training in the first stage.
\end{itemize}

\noindent\textbf{\textit{\\Testing description}}
They follow the official testing strategy \textit{without any aftertreatments.}
Following \cite{ pan2020ecg}, we compare our method with other evaluated methods on MVP  original dataset (2,048 points and 16,384 points),
the evaluated CD loss and F-score  are reported in Table \ref{tabc2:compair_others} and \ref{tabc2:compair_others_16384}. 
Our method outperforms other methods in terms of CD and F-score@1\%.



\subsection{Solution of Third Place}
\noindent{\textit{Learning Spherical Point Transformation for Point Cloud Completion}}

\noindent{\textit{\\Team Members: Junsheng Zhou$^\star$, Xin Wen$^\star$, Peng Xiang, Yu-Shen Liu$^\ddagger$, and Zhizhong Han}}

\noindent $^\star$ denotes equally contribution

\noindent $^\ddagger$ denotes corresponding author

\noindent\textbf{\textit{\\General Method Description}}

\noindent Inspired by PMPNet\cite{wen2021pmp}, which learns the point moving path between the incomplete and complete points, they design a novel deep neural network for point cloud completion by learning spherical point transformation (SPTNet). 

As shown in Fig. \ref {figc3:model}, they first randomly sample points on the standard sphere and generate a complete point cloud by moving each spherical point. As a result, the network learns a strict and unique correspondence on point-level and thus improves the quality of the predicted complete shape. Moreover, due to the uniformity of spherical points, the predicted complete shape generated by the network also retains the uniformity. In order to fine-tune the results, they added the RENet\cite{pan2021variational} at the end of the network. Different from other supervised\cite{wen2020point, xiang2021snowflakenet} or unsupervised\cite{wen2021cycle4completion} point cloud completion methods, SPTNet makes full use of the spherical 
distribution, so it achieves better results. And they learned a non-end-to-end point movement based RENet for further fine-tuning. In summary, the main contributions of their work are listed as follows:
\begin{itemize}
\item[(1)] They propose a novel network for point cloud completion, named SPTNet, which moves spherical points to generate a complete point cloud in high accuracy.
\item[(2)] They propose to generate complete shapes in a coarse-to-fine manner. After spherical points transformation, they apply RENet in both end-to-end and non-end-to-end manners, which justified their idea of point movement based shape generation for fine-tuning.
\item[(3)] They explore the feasibility of leveraging a transformer-based network to learn point-wise features in the encoder, which captures more local information between points.
\end{itemize}

\noindent\textbf{\textit{\\Training Description}}

\noindent They did not pre-train their network under any additional datasets or adopt any pre-trained models. And they did not do any enhancement to the data. In the training process, they adopt a two-stage training strategy. In the first stage, SPTNet with a generative RENet is trained end-to-end. And in the second stage, based on the results generated in the first stage, they train an additional point movement based RENet for each class, which is non-end-to-end. And they use Chamfer Distance as the loss function during training.  Adam optimizer is used for all networks with an initial learning rate of 0.0001, and the learning rate is multiplied by 0.7 every 40 epochs. The batch size is set to 16, and the total number of training epochs is set to 100.
They use PyTorch to implement the method. All the models are trained on a NVIDIA 2080Ti GPU with 11GB memory consumption. It takes about 24 hours for training and 45 minutes for testing. And there is no human effort required for implementation, training, and validation.

\noindent\textbf{\textit{\\Testing Description}}

\noindent During testing, the first pass the test data through SPTNet, and then input it into different point movement based RENet according to the category of the model.

\section{Partial-to-Partial Point Cloud Registration} \label{sec:reg}
\paragraph{Overview.}
Besides completion for single-view partial point clouds, researchers often perform Partial-to-Partial Registration (PPR) for 3D reconstruction.
However, previous methods usually perform PPR on uniformly dis ModelNet40~\cite{wu20153d} under restricted rotations in [0, 45\textdegree].
On the contrary, we use virtual-scanned partial point cloud pairs in the MVP registration challenge, and many partial point cloud pairs are under unrestricted rotations in [0, 180\textdegree].
These settings are more similar to observations in real applications, such as 6D pose estimation, which, however, challenges existing object-centric PPR methods.

\begin{table}[t]
\centering
\setlength{\tabcolsep}{5pt}
\caption{Top team results in MVP Registration Challenge 2021.}
\begin{tabular}{l|c|cccc}
\toprule[1.5pt]
\fontsize{7.5}{7.5}\selectfont Method & \fontsize{7.5}{7.5}\selectfont Ranking & \fontsize{7.5}{7.5}\selectfont Rot Error & \fontsize{7.5}{7.5}\selectfont Trans Error & \fontsize{7.5}{7.5}\selectfont MSE \\
\midrule\midrule
\fontsize{7.5}{7.5}\selectfont IM-Net & \fontsize{7}{7}\selectfont 3 & \fontsize{7}{7}\selectfont 2.91 & \fontsize{7}{7}\selectfont 0.027 & \fontsize{7}{7}\selectfont 0.078 \\
\fontsize{7.5}{7.5}\selectfont ROPNet + PREDATOR & \fontsize{7}{7}\selectfont 2 & \fontsize{7}{7}\selectfont 2.97 & \fontsize{7}{7}\selectfont 0.026 & \fontsize{7}{7}\selectfont 0.078 \\
\fontsize{7.5}{7.5}\selectfont Hybrid Optimization & \fontsize{7}{7}\selectfont 1 & \fontsize{7}{7}\selectfont 2.92 & \fontsize{7}{7}\selectfont 0.021 & \fontsize{7}{7}\selectfont 0.072 \\
\bottomrule[1.5pt]
\end{tabular}
\label{tab:all_reg}
\end{table}

\paragraph{Dataset.}
We generate partial point cloud pairs from the MVP dataset~\cite{pan2021variational}, and a successful pair is selected if sufficient overlapped areas are detected.
In total, we generate a training set with 6,400 paired partial point clouds, a test set with 1,200 pairs, and an extra-test set with 2,000 pairs.
In the test set and extra-test set, most relative rotations are within [0, 45\textdegree], and the rest have unrestricted rotations $\in$ [0, 180\textdegree]. The ratio is roughly 4 : 1.
Note that the source and the target are two different incomplete point clouds scanned from the same CAD model.

\paragraph{Evaluation Metric.}
We mainly use three metric, including isotropic rotation error $\mathcal{L}_\mathbf{R}$, l2 translation error $\mathcal{L}_\mathbf{t}$ and an MSE error $\mathcal{L}_{\text{MSE}}$ considering both rotations and translations.
Those metric functions are defined as:
\begin{equation}
    \mathcal{L}_\mathbf{R} = \text{rad2deg}\Big(\arccos\big(\frac{1}{2}(tr(\mathbf{R}_{\text{GT}}^{-1} \cdot \mathbf{R}_{\text{pred.}})-1)\big)\Big)\;,
\end{equation}
\begin{equation}
    \mathcal{L}_\mathbf{t} = \big\| \mathbf{t}_{\text{GT}} - \mathbf{{t}}_{\text{pred.}} \big\|_2\;,
\end{equation}
\begin{equation}
    \begin{split}
        \mathcal{L}_{\text{MSE}} = \mathcal{L}_\mathbf{t} +& \text{deg2rad}(\mathcal{L}_\mathbf{R})
        = \big\| \mathbf{t}_{\text{GT}} - \mathbf{{t}}_{\text{pred.}}\big\|_2 \\ +& \arccos\big(\frac{1}{2}(tr(\mathbf{R}_{\text{GT}}^{-1} \cdot \mathbf{R}_{\text{pred.}})-1)\big) \;,
    \end{split}
\end{equation}
where $\text{rad2deg}$ converts angles from radians to degrees, and $\text{deg2rad}$ is the inverse operation. 
$\mathbf{R}_{\text{GT}}$, $\mathbf{R}_{\text{pred.}}$, $\mathbf{t}_{\text{GT}}$ and $\mathbf{{t}}_{\text{pred.}}$ are groundtruth rotations, predicted rotations, groundtruth translations and predicted translations, respectively.

\paragraph{Results.}
The benchmark results are reported in Table~\ref{tab:all_reg}.

\subsection{Solution of First Place}
\noindent{\textit{Hybrid Optimization Method with Unconstrained Variables}}

\noindent\textit{\\Team Members: Yuanjie Yan and Junyi An}

\noindent\textbf{\textit{\\General Method Description}}

\noindent In the MVP registration, they introduces a new set of variables $[\mathbf{v}, \theta, \mathbf{u}, d]$ to directly represent the transformation from point cloud $\mathbf{P}$ and point cloud $\mathbf{Q}$, where $\mathbf{v}$, $\theta$, $\mathbf{u}$ and $d$ denote the rotation direction, rotation angle, translation direction and translation distance, respectively. They map restricted variables to unrestricted variables to optimize the translation vector.
Then, they use a variant of CD loss as optimization function. Finally, they initialize the variables and use strategies to optimize them.

\noindent\textbf{Transformation matrix with optimized variables:}
The rotation matrix $\mathbf{R}$ can be represented by rotation direction and angle.
\begin{equation}
    \begin{aligned}
        R &= \cos(\theta)\mathbf{I} + (1-\cos(\theta))\mathbf{v}\mathbf{v}^{T}+\sin(\theta)Skew(\mathbf{v}),\\
        \end{aligned}
\end{equation}
where 
\begin{equation}
        Skew(\mathbf{v}) =
        \begin{bmatrix}
            0&-v_{z}&v_{y}\\
            v_{z}&0&-v_{x}\\
            -v_{y}&v_{x}&0\\
        \end{bmatrix}.
\end{equation}
$\mathbf{v}$ can be any direction vector, and the range of $\theta$ is in the interval $[0,\pi]$. 

\begin{table}
\caption{The ablation experiments of the proposed method.}
\vspace{-5mm}
\setlength{\tabcolsep}{2.7pt}
\begin{center}
 \begin{tabular}{cccc|ccc}
   \toprule[1.5pt]
   \begin{tabular}[c]{@{}c@{}}\footnotesize Local \\ \footnotesize CD loss\end{tabular} & \begin{tabular}[c]{@{}c@{}}\footnotesize Projected \\ \footnotesize CD loss\end{tabular} & \begin{tabular}[c]{@{}c@{}}\footnotesize Unconstrained \\ \footnotesize T \end{tabular} & \footnotesize Strategies & \begin{tabular}[c]{@{}c@{}}\footnotesize Rot \\ \footnotesize Error\end{tabular} & \begin{tabular}[c]{@{}c@{}}\footnotesize Trans \\ \footnotesize Error\end{tabular} & \footnotesize MSE \\
   \midrule\midrule
    \checkmark & & \checkmark &\checkmark & \footnotesize 3.2654 & \footnotesize 0.0230 & \footnotesize 0.0799 \\
    \checkmark & \checkmark &  & \checkmark & \footnotesize 3.2350 & \footnotesize 0.0292 & \footnotesize 0.0796 \\
    \checkmark & \checkmark &  \checkmark &  & \footnotesize 9.8391 & \footnotesize 0.0239 & \footnotesize 0.2149 \\
   \checkmark & \checkmark & \checkmark & \checkmark & \footnotesize 2.9497 &  \footnotesize 0.0211 & \footnotesize 0.0726  \\
   \bottomrule[1.5pt]
\end{tabular}
\end{center}

\label{tabr1:tab1}
\end{table}

The translation vector $\mathbf{T}$ can be represented by translation direction and distance.
\begin{equation}
\mathbf{T} = d \mathbf{u}
\end{equation}
To ensure the legitimacy of the translation, they provide a constraint on the translation distance: $|\mathbf{T}|<=d_{max}$. 

The constraint with variables is known as "box constraint" in the optimization literature. There are three different methods of approaching this problem following \cite{carlini2017towards}:
\begin{itemize}
 \item[(1)] Projected gradient descent performs one step of standard gradient descent and then clips all the coordinates to be within the box.
 \item[(2)] Clipped gradient descent does not clip the original target on each iteration; rather, it incorporates the clipping into the objective function to be minimized. 
 \item[(3)] Change of variables introduces a new variable instead of optimizing the original variable. 
\end{itemize}

They choose the third approach as a smoothing of clipped gradient descent that eliminates the problem of getting stuck in extreme regions. In this case, the translation vector is given by:
\begin{equation}
\mathbf{T} = d_{max}Sigmoid(d_{u})\mathbf{u}.
\end{equation}
where the constrained $d$ is replaced by term $d_{u}$. For similar ideas, they also proposed a mapping method using the $\sin$ function.
\begin{equation}
\begin{aligned}
\mathbf{T} = 1/2(d_{max} + d_{max}Sin(d_{u}))\mathbf{u}.
\end{aligned}
\label{eq5}
\end{equation}
They try several mapping functions. The results show that eq.~\eqref{eq5} is the most effective mapping to reduce the Trans error.

\noindent\textbf{Loss:}
To optimize the variables, they introduce the chamfer distance (CD) loss, which can help to find the point correspondences in the optimization process. The standard CD loss is given by:
\begin{equation}
\begin{aligned}
\mathcal{L}_{C D}(\mathbf{P}, \mathbf{Q}) &=\frac{1}{|\mathbf{P}|} \sum_{x \in \mathbf{P}} \min _{y \in \mathbf{Q}}\|x-y\|^{2} \\
&+\frac{1}{|\mathbf{Q}|} \sum_{y \in \mathbf{Q}} \min _{x \in \mathbf{P}}\|x-y\|^{2}.
\end{aligned}
\end{equation}

Moreover, they propose two variants of CD loss to improve the registration of MVP, \textit{Local CD Loss} $\mathcal{L}_{C D}^{local}$ and \textit{Projected CD Loss} $\mathcal{L}_{C D}^{uv}$.

\begin{table}
    \caption{Registration comparison between different methods.}\label{tabr1:tab2}
    \vspace{-5mm}
    \small
    \begin{center}
    \begin{tabular}{l|ccc}
    \toprule[1.5pt]
    Method & Rot Error & Trans Error & MSE \\
    \midrule\midrule
    DCP & 27.4702 & 0.1042 & 0.5836 \\
    IDAM & 20.6607 & 0.1482 & 0.5088 \\
    Euler angles & 13.7002 & 0.0488 & 0.2879 \\
	6-D &  9.6956 & 0.0442 & 0.2134 \\
	\hline
    \rowcolor{Gray} Hybrid Optimization & 2.9187 & 0.0206 & 0.0716\\
    \bottomrule[1.5pt]
    \end{tabular}
    \end{center}
\end{table}

\begin{itemize}
    \item[$\bullet$] \textbf{Local CD loss}. 
    The local CD loss on point clouds reduces the limit of matching all points, improving the tolerance for point cloud distribution differences.
    
    \begin{equation}
        \begin{aligned}
            \mathcal{L}_{C D}^{local}(\mathbf{P}, \mathbf{Q}, \alpha) &=\frac{1}{\left|\mathbf{P}_{\alpha}\right|} \sum_{x \in \mathbf{P}_{\alpha}} \min _{y \in \mathbf{Q}}\|x-y\|^{2} \\
            &+\frac{1}{\left|\mathbf{Q}_{\alpha}\right|} \sum_{y \in \mathbf{Q}_{\alpha}} \min _{x \in \mathbf{P}}\|x-y\|^{2},
        \end{aligned}
    \end{equation}
    where $\alpha$ is a hyperparameter between $0$ and $1$. $\mathbf{P}_\alpha$ and $\mathbf{Q}_\alpha$ are subsets of $\mathbf{P}$ and $\mathbf{Q}$, $|P_\alpha| = \alpha * |\mathbf{P}|$ and $|\mathbf{Q}_\alpha| = \alpha * |\mathbf{Q}|$.
    
    \item[$\bullet$] \textbf{Projected CD loss}. 
    The local CD loss only pays attention to the local alignment but ignores the global point cloud matching. They project $(\mathbf{P}_{i}, \mathbf{Q}_{i})$ onto the $x - y$, $y - z$, $x - z$ plane and calculate the projected CD loss as follows.
    
    \begin{equation}
        \begin{aligned}
            \mathcal{L}_{C D}^{uv}(\mathbf{P}, \mathbf{Q}) &=\frac{1}{\left|\mathbf{P}\right|} \sum_{x \in \mathbf{P}_{\alpha}} \min _{y \in \mathbf{Q}}\|x_{uv}-y_{uv}\|^{2} \\
            &+\frac{1}{\left|\mathbf{Q}\right|} \sum_{y \in \mathbf{Q}} \min _{x \in \mathbf{P}}\|x_{uv}-y_{uv}\|^{2},
        \end{aligned}
    \end{equation}
    where $uv$ is a projected plane.
\end{itemize}

The final optimization function is given by:
\begin{equation}
    \begin{aligned}
        \mathcal{L}(\mathbf{P},\mathbf{Q}) = \mathcal{L}_{C D}^{local}&(\mathbf{P},\mathbf{Q}, \alpha) 
        							+ \beta \Big(\mathcal{L}_{C D}^{xy}(\mathbf{P},\mathbf{Q}) \\
        							&+\; \mathcal{L}_{C D}^{yz}(\mathbf{P},\mathbf{Q})
        							+ \mathcal{L}_{C D}^{xz}(\mathbf{P},\mathbf{Q})\Big),
    \end{aligned}
\end{equation}
where $\beta$ is balance weights.

\begin{table}[]
\caption{The first line: Results of repeated experiments. The second line: The result of the proposed method with eight intervals.}
\vspace{-5mm}
\setlength{\tabcolsep}{5pt}
\small
\begin{center}
\begin{tabular}{ c | c | c }
\toprule[1.5pt]
Rot Error &  Trans Error &  MSE \\
\midrule\midrule
\small 2.9751 $\pm$ 0.0254 & \small 0.02106 $\pm$ 0.0003 & \small 0.07299 $\pm$ 0.0004 \\
\small 2.9187* & \small 0.0206* & \small 0.0716*  \\

\bottomrule[1.5pt]
\end{tabular}
\end{center}

\end{table}

\noindent\textbf{\textit{\\Training description}}

\noindent The proposed method does not require training.

\noindent\textbf{\textit{\\Testing Description}}

\noindent Predicting rigid transformation by gradient descent is a non-convex optimization problem. Therefore, it is crucial for optimization from multiple initializations. 
In their method, they initialize the term $\mathbf{v}$ into $64$ rotation directions. As for the initial angle, they make $64$ initial value in the interval $[0, \pi / 4]$. The result with the smallest loss is selected as the rotation matrix. They set a threshold for the optimization loss. When the optimization loss is bigger than the threshold, they will initialize the angle in the intervals $[\pi / 4, \pi / 2]$, $[\pi / 2, 3 \pi  / 4]$, and $[3 \pi / 4, \pi ]$, respectively, and repeat the optimization. This optimization strategy divided intervals tends to solve the problem of the symmetric point cloud model. It tries to avoid the situation of rotating more than $\pi/ 4$ or $\pi / 2$. This strategy coincides with the prior probability of the dataset.
Most relative rotations are restricted in $[0, \pi / 4]$, and the rest have unrestricted rotations $\in$ $[0, \pi]$. The ratio is roughly $4 : 1$.

\begin{figure}
    \centering
    \includegraphics[width=1\linewidth]{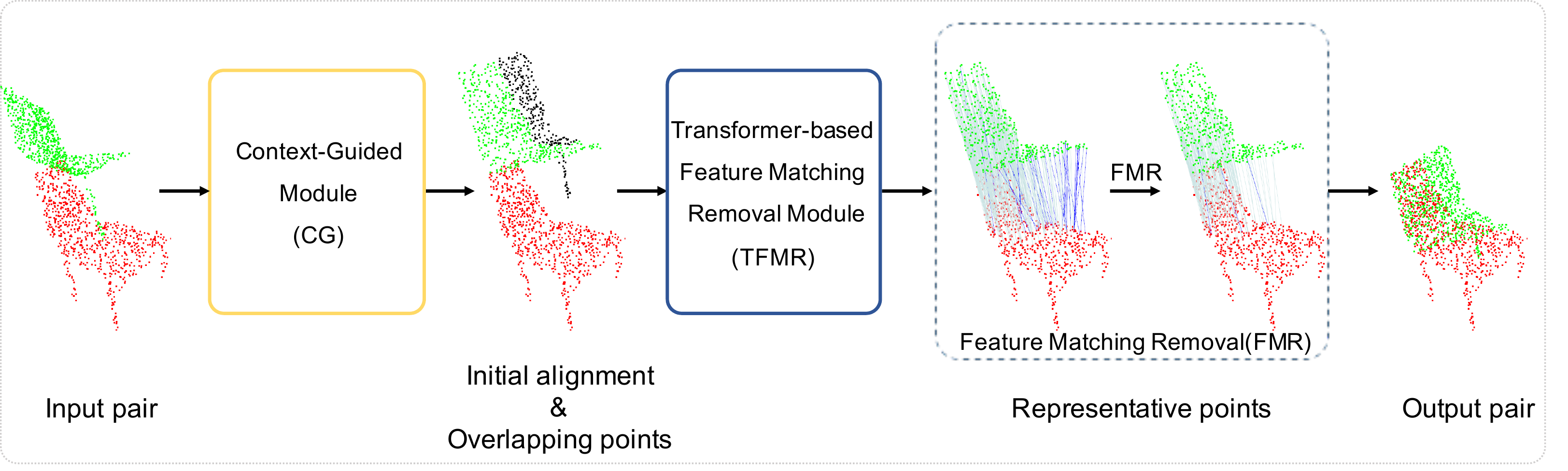}
    \caption{Overview of ROPNet registration pipeline. The CG module consumes the source (green) and target (red) point clouds, and outputs initial pose and overlapping points (non-overlapping points are in black). The TFMR module takes the output of CG module as input, and generates accurate correspondences.
    The FMR step removes false correspondences (blue lines) and keeps some positive correspondences (gray lines).}
    \label{figr2:ROPNet}
\end{figure}
\begin{figure}[t]
    \begin{center}
       \includegraphics[width=1\linewidth]{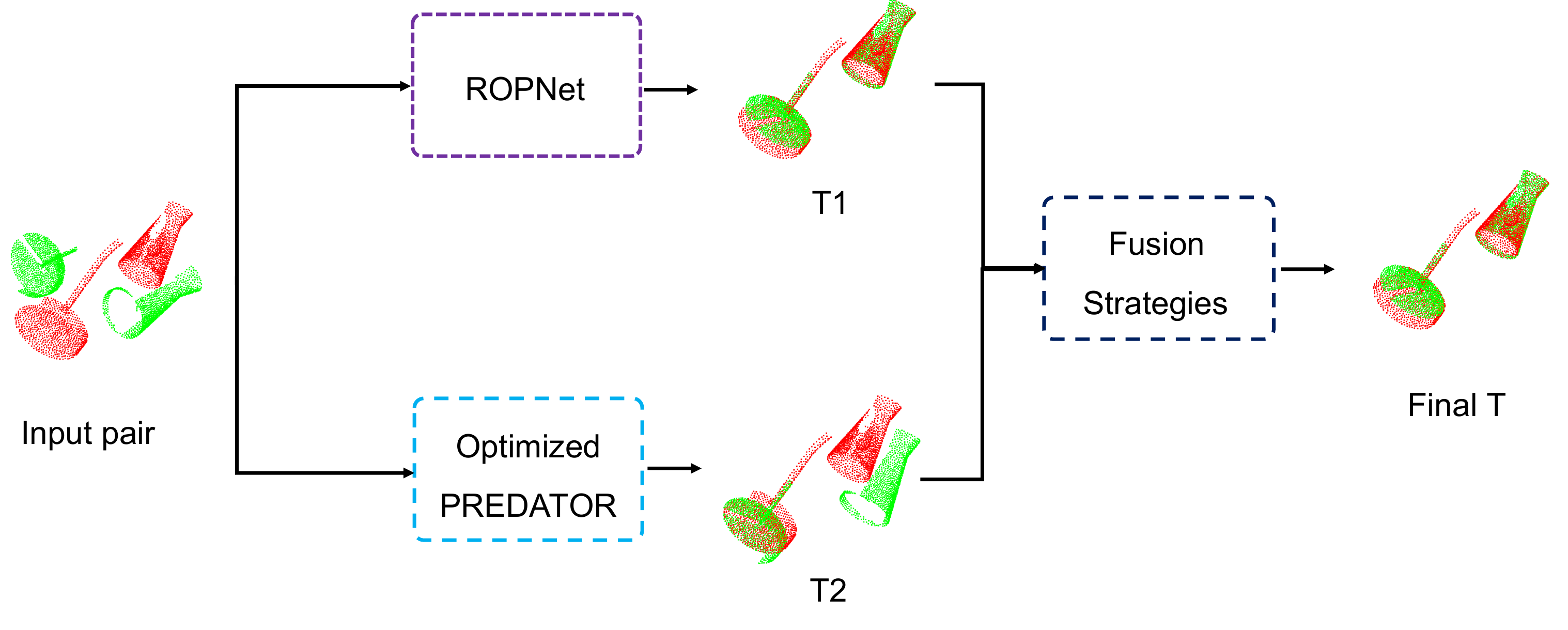}
    \end{center}
    \vspace{-4mm}
    \caption{The pipeline of our solution to the MVP Registration Challenge. The point cloud in green is the source point cloud,  and the point cloud in red is the target point cloud.}
\label{figr2:pipeline}
\end{figure}
\subsection{Solution of Second Place}
\noindent{\textit{Deep Models with Fusion Strategies}}

\noindent{\textit{\\Team Members: Lifa Zhu, Changwei Lin, Dongrui Liu, Xin Li, Francisco Gómez-Fernández.}}

\begin{figure*}[t]
\centering
\includegraphics[width=0.8\linewidth]{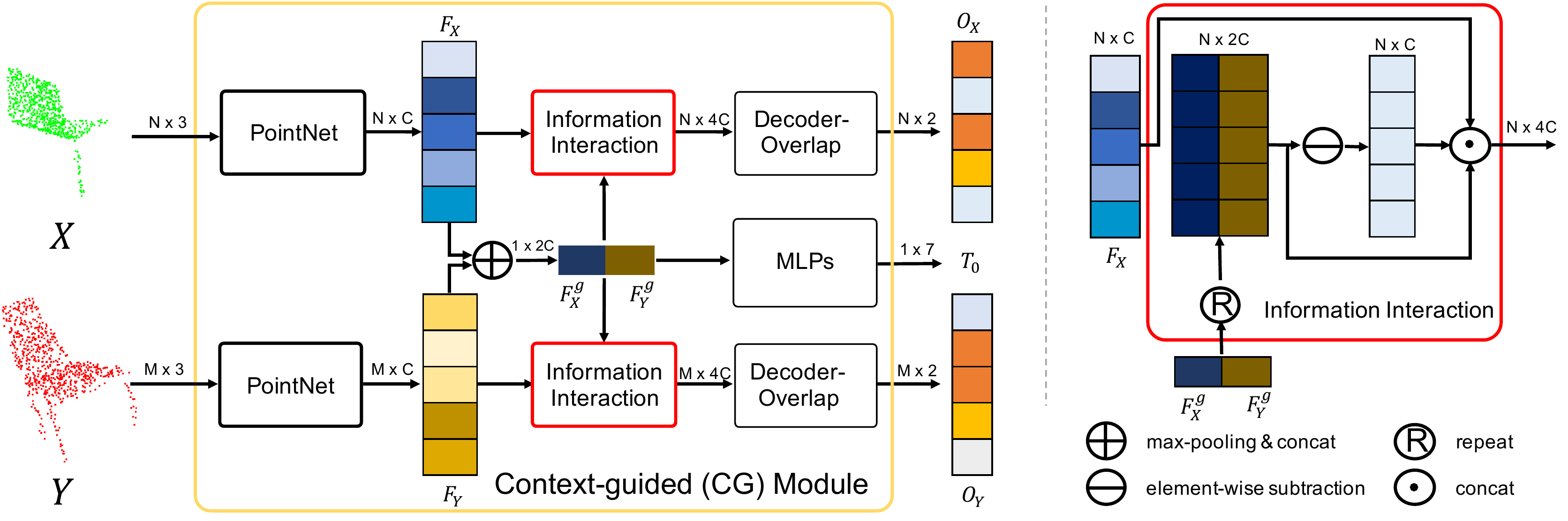} \\
\caption{Left: Architecture of the CG module. CG module consumes source $X$ (in green) and target $Y$ (in red) data, and outputs overlap score ($O_X$, $O_Y$) and initial transformation matrix $T_{0}$. Right: Details of information interaction. It takes point features and global features as input and outputs fused point features based on the pair.}
\label{figr2:ROP_CG}
\end{figure*}
\begin{figure*}[t]
\centering
\includegraphics[width=0.8\linewidth]{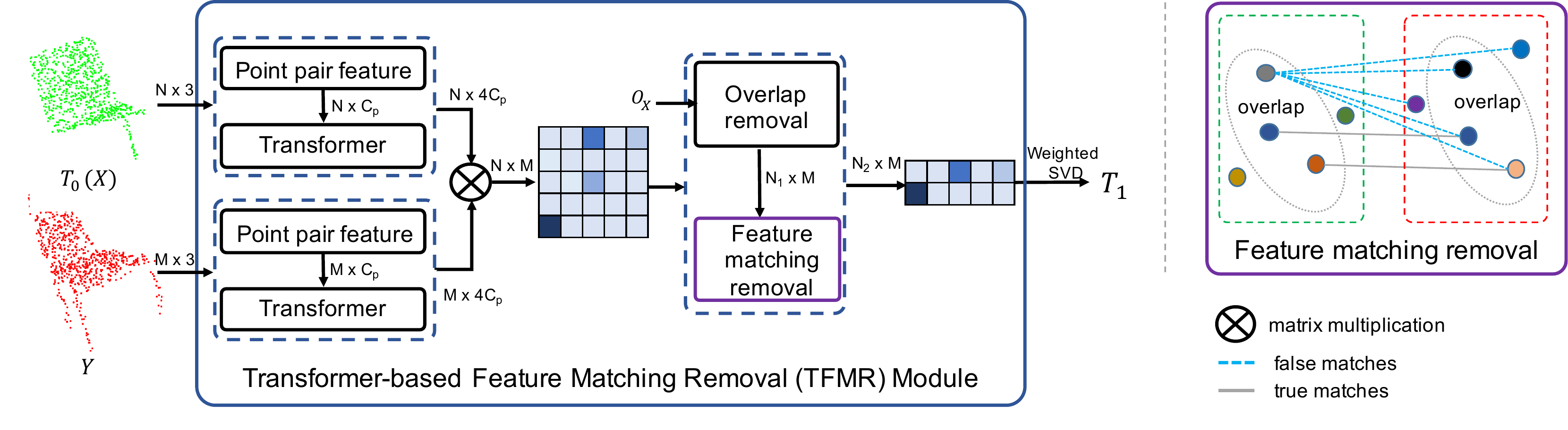} \\
\vspace{-1mm}
\caption{Left: Overview of Transformer-based feature matching removal (TFMR) module. TFMR module takes the transformed source $X'$ and target $Y$ as input, and outputs representative points and their correspondences. $T_0$ and $O_X$ are the output from CG module that denote initial alignment and overlap score for $X$.
Right: Details of feature matching removal (FMR). It takes correspondences for overlapping source points and outputs accurate correspondences (gray lines).}
\label{figr2:ROP_TFMR}
\end{figure*}

\noindent\textbf{\textit{\\General Method Description}}

\noindent They propose to fuse ROPNet~\cite{zhu2021point} and PREDATOR\cite{huang2021predator} to solve point cloud registration in the MVP Challenge. ROPNet and the pipeline of their solution to this challenge can be seen in Fig.~\ref{figr2:ROPNet} and  Fig.~\ref{figr2:pipeline}.

In ROPNet, they proposed a context-guided module that uses an encoder to extract global features for predicting point overlap score and introduced a Transformer to enrich point features and remove non-representative points based on point overlap score and feature matching. 

Considering unrestricted rotations, they used PREDATOR in their pipeline.
In PREDATOR source code, they found and solved a simple but important GNN bug that helps the network to obtain a higher performance.
Then, they adjusted parameters in PREDATOR for the MVP registration challenge. Inspired by the idea of partial-to-complete proposed in ROPNet, they try to remove some points in source data based on the predicted scores and keep all points in target data during RANSAC iterations.

Based on the above discussions, they designed a few ensemble strategies based on data characteristics to help fuse ROPNet and PREDATOR to estimate the final rigid transformation. 
Experiments on the test set showed it is effective in most cases, with few fails. 
ROPNet can be seen in Fig.~\ref{figr2:ROPNet}.
Their proposed context-guided (CG) module and Transformer-based Feature Matching Removal (TFMR) Module in ROPNet can be seen in Fig.~\ref{figr2:ROP_CG} and Fig.~\ref{figr2:ROP_TFMR}.
The pipeline of their solution is shown in Fig.~\ref{figr2:pipeline}.

\begin{table}
    \caption{Comparison to other approaches on val set.}
    \vspace{-7mm}
    \begin{center}
    \scalebox{0.7}{
        \begin{tabular}{l|c|c|c|c}
            \toprule[1.5pt]
            Model & rot\_level & Rot Error & Trans Error & MSE  \\
            \midrule\midrule
            RPMNet\_corr~\cite{zodage2020correspondence} & 0 & 12.5560 & 0.1674 & 0.3865 \\
            ROPNet~\cite{zhu2021point} & 0 & 1.0449 & 0.0193 & 0.0375 \\
            RPMNet\_corr~\cite{zodage2020correspondence} & 0, 1 & 21.9685 & 0.2062 & 0.5896 \\
            \hline
            \rowcolor{Gray} ROPNet~\cite{zhu2021point} + PREDATOR~\cite{huang2021predator} &  0, 1 & 3.1666 & 0.0292 & 0.0845 \\
            \bottomrule[1.5pt]
        \end{tabular}}
    \end{center}
    
    \label{tabr2:other_method}
\end{table}
\begin{figure*}[]
    \centering
    \includegraphics[width=0.8\linewidth]{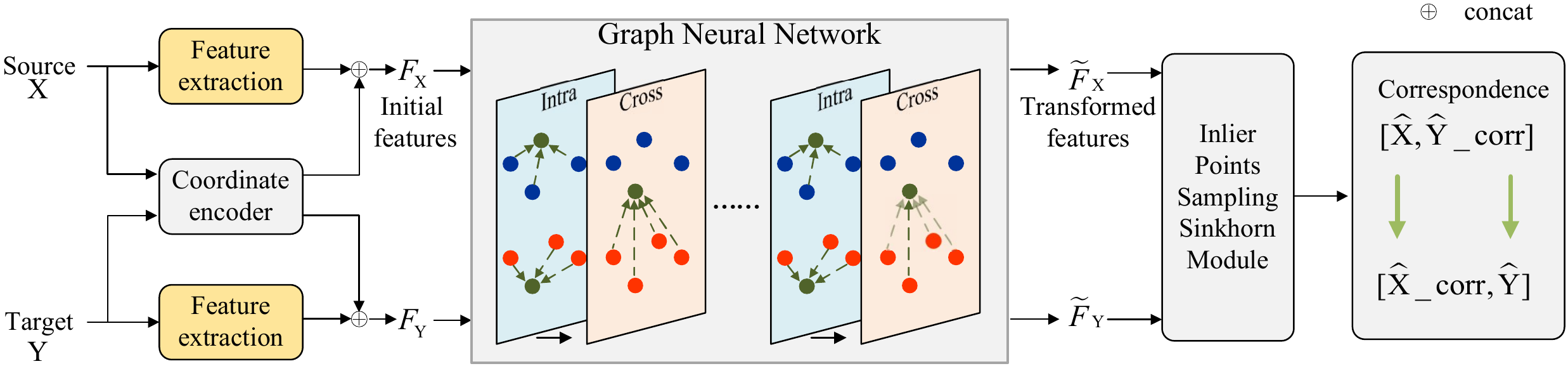}
    \vspace{-1mm}
    \caption{Overview of the proposed method. Feature extraction is a shared module. $\widehat{X}$ and $\widehat{Y}$ represent point clouds after sampling.}
    \label{figr3:overview}
\end{figure*}

\noindent\textbf{\textit{\\Training description}}

\noindent They trained ROPNet and PREDATOR independently. All 2048 points were involved in training for the two models.
For ROPNet, they trained for 600 epochs using Adam optimizer with an initial learning rate of 0.0001. The learning rate changes using a cosine annealing schedule. 
They trained ROPNet in a non-iterative manner. 
However, they run 2 iterations for the TFMR module during the test.
It is noted that they only trained ROPNet for small rotation angles ranging from 0° to 45°. 
Also, they did not use Point Pair Features \cite{drost2010model}, because they could not get accurate normal vectors in the MVP challenge data. 

For PREDATOR, following the code\footnote{\url{https://github.com/overlappredator/OverlapPredator}} released by the authors, they trained on MVP registration dataset for 200 epochs using SGD with 0.98 momentum. The initial learning rate was 0.01, with an exponential decay factor of 0.95 every epoch.
They trained PREDATOR under unrestricted rotation angles ranging from 0° to 360°. In addition, they adjusted some parameters such as voxel size to 0.04, sampled points in circle loss, and others in loss implementation.

\noindent\textbf{\textit{\\Testing Description}}

\noindent For each source and target point cloud pair, they estimate transformations $T_1 \in \mathbf{SE(3)}$ and $T_3 \in \mathbf{SE(3)}$ based on ROPNet and PREDATOR, respectively. $T_1$ is the output predicted from source to target using the end-to-end ROPNet model. $T_3$ is also the transformation from source to target, which is obtained with RANSAC using features and keypoints provided by PREDATOR. They select $T1$ or $T_3$ based on their proposed ensemble rules.
They compare their method with  RPMNet\_corr~\cite{zodage2020correspondence}, which is the variant of RPMNet~\cite{yew2020rpm} to help solve registration with unrestricted rotations. 
The results in Table~\ref{tabr2:other_method} show that their ROPNet achieves a much lower registration error than RPMNet\_corr when rot\_level is 0. When rot\_level is not restricted, the ensemble model of ROPNet and PREDATOR also achieves much lower registration error than RPMNet\_corr.

\subsection{Solution of Third Place}
\noindent{\textit{IM-Net: Partial Point Cloud Registration with Inliers Prediction and Matching}}

\noindent{\textit{\\Team Members: Qinlong Wang and Yang Yang}}

\begin{figure*}[]
    \centering
    \includegraphics[width=0.8\linewidth]{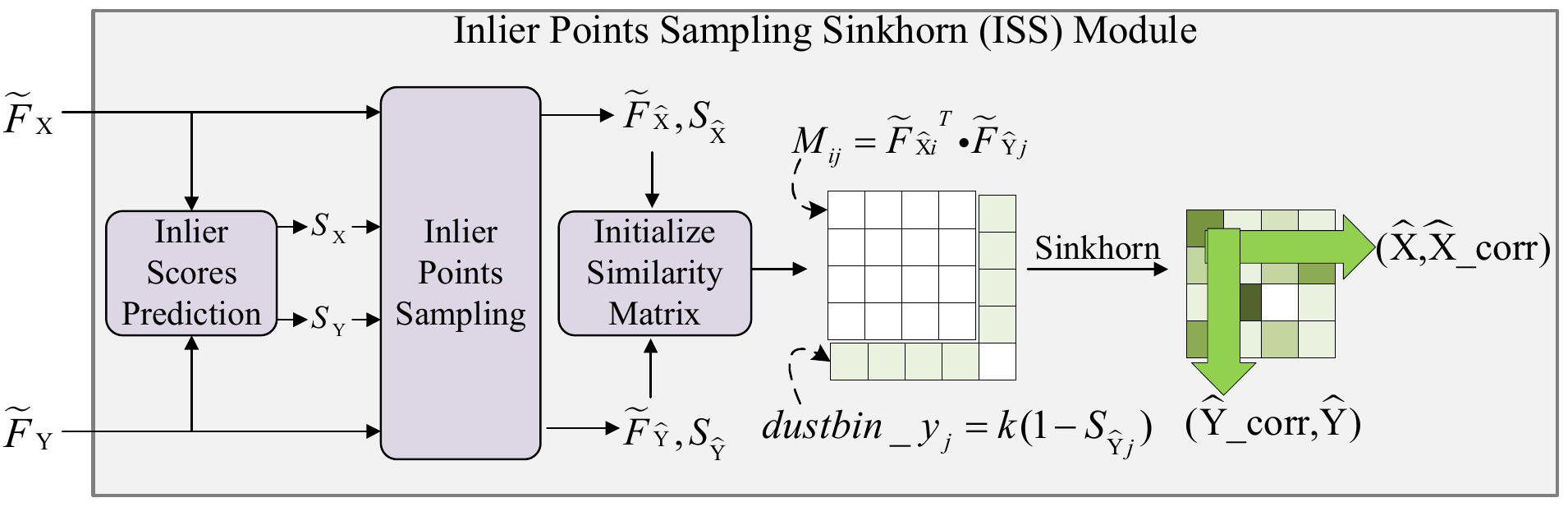}
    \caption{ISS module details. $k$ is a learnable parameter and shared for all points.}
    \label{figr3:ISS}
\end{figure*}
\noindent\textbf{\textit{\\General Method Description}}

\noindent Their work titled with, IM-Net: Partial Point Cloud Registration with \textbf{I}nliers Prediction and \textbf{M}atching is built with 4 main components, as illustrated in Figure \ref{figr3:overview}. 

\begin{itemize}
    \item[(1)] Firstly, they use DGCNN \cite{wang2019dynamic} to extract the per point initial feature in each point cloud. 
    \item[(2)] To extract identifying matching features, each point feature of two point cloud is then transformed via a graph neural network constructed by interleaving intra- and cross-graph aggregations, i.e., edges of intra-graph connect all points in the same point cloud and edges of cross-graph connect per point with all points in another point cloud. They introduce a fully connected, equal-weighted intra-graph to learn the relative structure information of each point cloud, in which a large receptive field is beneficial to clarifying ambiguous features. And an attention-based cross-graph is adopted to communicate information between two point clouds for identifying point features. To better learn the structure information, they embed the coordinate of each point with MLP to the same dimension with the initial feature and then concatenate two features before feeding into GNN. 
    \item[(3)] Then an Inlier points Sampling Sinkhorn (ISS) module samples and matches confidential inlier points in overlapping parts. Firstly, inlier scores for all points are predicted to prepare sampling candidates. Especially in the training phase, they sample half of the points with higher scores and half of the points with lower scores. And in the test phase, the sample is simply to select all candidates with higher scores. Then, they construct a new similarity matrix with dustbin by sampling features. To further handle outliers lain in sampled points, the dustbin score is initialized with inlier scores. Next, they feed the similarity matrix into Sinkhorn algorithm to generate a matching confidence map. 
    \item[(4)] Finally, a bidirectional correspondence is constructed by concatenating source to target correspondence, and the symmetric one, i.e. correspondence, is selected by the matches with the highest confidence in each row and column of the matching confidence map. The introduced correspondence can generate more matching pairs for low overlapping data and show better results than the correspondence based on mutual check. Each correspondence is weighted by the corresponding confidence. They compute the transformation using SVD. 
\end{itemize}

\noindent\textbf{Model Details:} For the submitted model, they interleave the intra- and cross-graph for 10 layers. The dimension of the initial features extracted by DGCNN \cite{wang2019dynamic} is 64, and the feature dimension in GNN and ISS module remains at 128 after concatenation. The ISS module will sample 1/6 points for each point cloud. They use the same evaluation metrics with the official implementation of MVP benchmark.

\begin{table}[]
\setlength{\tabcolsep}{3.75pt}
\caption{Performance on MVP test data with Gaussian noise and ModelNet40.}
\vspace{-3mm}
\centering
\begin{tabular}{c|c|cccc|c}
    \toprule[1.5pt]
    \multirow{2}{*}{\fontsize{7.75}{7.75}\selectfont Dataset} & \multirow{2}{*}{\fontsize{7.75}{7.75}\selectfont Preprocess} & \multicolumn{4}{c|}{\fontsize{7.75}{7.75}\selectfont Error} & \multirow{2}{*}{\fontsize{7.75}{7.75}\selectfont Recall}\\
    \cmidrule(lr){3-6} 
     & & \fontsize{7}{7}\selectfont Rot. & \fontsize{7}{7}\selectfont Trans. & \fontsize{7}{7}\selectfont MSE & \fontsize{7}{7}\selectfont RMSE &   \\
    \midrule\midrule
    \fontsize{7}{7}\selectfont MVP test~\cite{pan2021robust} & - & \fontsize{6}{6}\selectfont 2.5008 & \fontsize{6}{6}\selectfont 0.0305 & \fontsize{6}{6}\selectfont 0.0742 & \fontsize{6}{6}\selectfont 0.0382 & \fontsize{6}{6}\selectfont 0.9583 \\
    \fontsize{7}{7}\selectfont MVP test~\cite{pan2021robust} & \fontsize{7}{7}\selectfont add noise & \fontsize{6}{6}\selectfont 3.2183 & \fontsize{6}{6}\selectfont 0.0311 & \fontsize{6}{6}\selectfont 0.0872 & \fontsize{6}{6}\selectfont 0.0447 & \fontsize{6}{6}\selectfont 0.9633 \\
    \fontsize{7}{7}\selectfont ModelNet40 \cite{wu20153d} & \fontsize{7}{7}\selectfont sample rotation & \fontsize{6}{6}\selectfont 3.7722 & \fontsize{6}{6}\selectfont 0.0120 & \fontsize{6}{6}\selectfont 0.0779 & \fontsize{6}{6}\selectfont 0.0236 & \fontsize{6}{6}\selectfont 0.9617 \\
    \bottomrule[1.5pt]
    \end{tabular}
    \label{tabr3:1}
\end{table}

\noindent\textbf{Implementation Details:} The loss function used for training consists of 4 parts. The first two parts are a cross-entropy loss for inlier scores and a peaky loss inspired by the loss function in R2D2 \cite{revaud2019r2d2} to maximize the peakiness of the overlap scores. The last two parts are a cross-entropy loss for match confidence map following the implementation of the loss function for RPMNet in \cite{zodage2020correspondence} and a cross-entropy loss for dustbin confidence. They weighted the peaky loss with 0.25 and the dustbin loss with 0.5.

\noindent\textbf{\textit{\\Training description}}

\noindent They train their network using MVP training data and validate the model with MVP test data. During training, they random sample each point cloud to 1024 points for saving memory and augmenting the overlap rate of data. Moreover, the ISS module will sample half of the positive samples and half of the negative samples, i.e. matching and unmatching points. The network is trained using ADAM optimizer \cite{kingma2014adam} with a learning rate of 0.0005. The submitted model is refined within 500 epochs using a learning rate of 0.0001 after 1000 epochs. The network often converges after 500 epochs.

\noindent\textbf{\textit{\\Testing Description}}

\noindent During the test time, they remain 2048 points for inference, and ISS module will sample points with higher inlier scores. They further select half of the bidirectional correspondence with higher correspondence confidence to compute the transformation.

\section{Discussion}

\subsection{Completion}
\noindent\textbf{Summary.}
Generally, the top-3 methods significantly outperform baseline methods.
PoinTr++ employs a global transformer for generating the missing parts, followed by using RENet.
CRNet proposes the IOI augmentation and multi-scale SPD module to achieve semantic-aware completion.
SPTNet makes full use of the spherical distribution, and it learns a non- end-to-end point movement based RENet for further fine-tuning.

\noindent\textbf{Future Directions.}
In this challenge, we use CD loss for evaluating different completion methods.
However, CD loss is not sensitive to global distribution, and the completion results can be biased to partial observations.
New metrics, such as BCD~\cite{wu2021density} and EMD, can be leveraged for evaluation.
Recently, diffusion models~\cite{lyu2021conditional} provide impressive point cloud completion results, which also circumvent the imbalance issue.
In addition, unsupervised~\cite{zhang2021unsupervised} or self-supervised~\cite{wen2021cycle4completion} point cloud completion can also be studied.

\subsection{Registration}
\noindent\textbf{Summary.}
In a nutshell, the top-3 methods achieve surprisingly good registration results for partial-to-partial point cloud registration, especially for those with unrestricted rotations.
The 1st place method uses a non-learning hybrid optimization.
The 2nd place uses an assembling strategy with ROP~\cite{zhu2021point} and PREDATOR~\cite{huang2021predator}.
The 3rd place uses cross-graph connections and ISS module.

\noindent\textbf{Future Directions.}
Although they achieve outstanding registration results, the PPR problem has not been fully resolved, as:
1) hybrid optimization requires multiple initializations;
2) the assembling method is not elegant or efficient;
3) the 3rd place method
heavily relies on cross-graph connections and 
requires overall 1500 epochs for training.
Moreover, those registration methods did not take full advantage of pose-invariant features (\eg \textit{PPF}), which can facilitate full-range PPR~\cite{pan2021robust}.

\section*{Acknowledgement}
\noindent We sincerely thank Yuanhan Zhang for helpful discussions.

{\small
\bibliographystyle{ieee_fullname}
\bibliography{egbib}
}

\end{document}